%% file: main.tex
\documentclass[journal]{IEEEtran}
\IEEEoverridecommandlockouts
\usepackage{cite}
\usepackage{amsmath,amssymb,amsfonts}
\usepackage{amsthm}
\usepackage{relsize}
\usepackage{tabularx}
\usepackage{subfig}
\usepackage{url}
\usepackage{textcomp}
\usepackage{xcolor}
\usepackage{nicefrac}       
\usepackage[pdftex]{graphicx}
\usepackage{multirow}
\usepackage{xspace}
\usepackage{enumitem}
\usepackage{booktabs}
\usepackage{tabu}
\usepackage{siunitx}
\usepackage{algorithm}
\usepackage{algpseudocode}
\usepackage[english]{babel}

\newcommand{\algo}{CTAB-GAN\xspace}
\newcommand{\algoplus}{CTAB-GAN+\xspace}
\newcommand{\ctgan}{CTGAN\xspace}
\newcommand{\tablegan}{TableGAN\xspace}
\newcommand{\cwgan}{CWGAN\xspace}
\newcommand{\medgan}{MedGAN\xspace}
\newcommand{\encoder}{Mixed-type Encoder\xspace}
\newcommand{\imbalanced}{imbalanced\xspace}
\newcommand{\Imbalanced}{Imbalanced\xspace}
\newcommand{\skewed}{skewed\xspace}
\newcommand{\Skewed}{Skewed\xspace}
\newcommand{\column}{variable\xspace}
\newcommand{\columns}{variables\xspace}
\newcommand{\variable}{variable\xspace}
\newcommand{\variables}{variables\xspace}
\newcommand{\dsm}{downstream\xspace}
\newcommand{\Dsm}{Downstream\xspace}
\newcommand{\wgp}{Was+GP\xspace}

\newcommand{\bimodal}{mixed\xspace}
\newcommand{\mixed}{mixed\xspace}
\newcommand{\Mixed}{Mixed\xspace}

\newtheorem{cor}{\textbf{Corollary}}
\newtheorem{prf}{\textbf{Proof}}

\newif\ifshowcomments
\showcommentstrue
 
\ifshowcomments 
\newcommand{\mynote}[2]{\fbox{\bfseries\sffamily\scriptsize{#1}}
{\small$\blacktriangleright$\textsf{#2}$\blacktriangleleft$}}
\else
\newcommand{\mynote}[2]{}
\fi

\newcolumntype{L}[1]{>{\raggedright\let\newline\\\arraybackslash\hspace{0pt}}m{#1}}
\newcolumntype{C}[1]{>{\centering\let\newline\\\arraybackslash\hspace{0pt}}m{#1}}
\newcolumntype{R}[1]{>{\raggedleft\let\newline\\\arraybackslash\hspace{0pt}}m{#1}}

\newif\ifshowntationtable
\showntationtablefalse

\author{

   \IEEEauthorblockN{Zilong Zhao\IEEEauthorrefmark{1}\textsuperscript{\textsection}
   ,  Aditya Kunar\IEEEauthorrefmark{1}\textsuperscript{\textsection}, Robert Birke\IEEEauthorrefmark{2}, Lydia Y. Chen
    \IEEEauthorrefmark{1} }
    \\
    \IEEEauthorblockA{\IEEEauthorrefmark{1}TU Delft, Netherlands
    \{z.zhao-8, a.kunar, y.chen-10\}@tudelft.nl}
    \\
    \IEEEauthorblockA{\IEEEauthorrefmark{2}ABB Research, Switzerland
    \{robert.birke\}@ch.abb.com}
}
\title{\algoplus: Enhancing Tabular Data Synthesis}

\begin{document}

\maketitle
\begingroup\renewcommand\thefootnote{\textsection}
\footnotetext{Equal contribution}
\endgroup

\input{abstract}
\input{introduction_new}
\input{motivation_new}
\input{background_new}
\input{model_v2_new}
\input{experiment_new}
\input{experiment_dp}
\section{Conclusion}

Motivated by the importance of data sharing and fulfillment of governmental regulations, we propose \algoplus, a conditional GAN based tabular data generator. \algoplus advances beyond SOTA methods by modeling \bimodal \variables and providing strong generation capability for imbalanced categorical \variables, 
and continuous \variables with complex distributions.
The core features of \algoplus include: (1) introduction of the auxiliary component, i.e., classifier or regressor, into conditional GAN, (2) effective data encodings for \bimodal and simple Guassian \variables, (3) a novel construction of conditional vectors and (4) tailored DP discriminator for tabular GAN. We exhaustively evaluate \algoplus against state-of-the-art baselines on seven  tabular datasets under a wide range of metrics, namely 
for ML utility and statistical similarity. Results show that the synthetic data of \algoplus results into higher ML utility and higher similarity. The overall improvement on classification datasets is at least 56.4\% (AUC) and 41.2\% (accuracy) compared to related work with no privacy guarantees. When turning on differential privacy, \algoplus averagely outperforms at least 48.16\% on accuracy and 38.05\% on F1-Score than existing DP-GAN training under all privacy budgets. 
The substantial results of \algoplus demonstrate its potential for a wide range of applications that greatly benefit from data sharing, such as banking, insurance, and manufacturing.






\bibliographystyle{abbrv}
\bibliography{acml21}






\end{document}

%% file: abstract.tex
\begin{abstract}

While data sharing is crucial for knowledge development, privacy concerns and strict regulation (e.g., European General Data Protection Regulation (GDPR)) limit its full effectiveness. 
Synthetic tabular data emerges as alternative to enable data sharing while fulfilling regulatory and privacy constraints. 
State-of-the-art tabular data synthesizers draw methodologies from Generative Adversarial Networks (GAN).
As GANs improve the synthesized data increasingly resemble the real data risking to leak privacy.
Differential privacy (DP) provides theoretical guarantees on privacy loss but degrades data utility. Striking the best trade-off remains yet a challenging research question.

We propose \algoplus
a novel conditional tabular GAN. 
\algoplus improves upon state-of-the-art 
by 
(i) adding downstream losses to conditional GANs for higher utility synthetic data in both classification and regression domains;
(ii) using 
Wasserstein loss with gradient penalty for better training convergence; (iii) introducing novel encoders targeting mixed continuous-categorical \variables and \variables with unbalanced or skewed data; 
and (iv) training with DP stochastic gradient descent to impose strict privacy guarantees.
We extensively evaluate \algoplus 
on data similarity and analysis utility against state-of-the-art tabular GANs. 
The results show that \algoplus synthesizes privacy-preserving data 
with at least 48.16\% higher utility across multiple datasets and learning tasks under different privacy budgets.



\end{abstract}
\begin{IEEEkeywords}
GAN; Data synthesis; Tabular data; Differential privacy; Imbalanced distribution
\end{IEEEkeywords}

%% file: introduction_new.tex
\section{Introduction}
Many companies nowadays discover valuable 
insights from various internal and external data sources. However, the deep knowledge behind big data often violates personal privacy and leads to an unjustified analysis~\cite{narayanan2008}.  To prevent the abuse of data and the risks of privacy breaching, the European Commission introduced the European General Data Protection Regulation (GDPR)  and enforced strict data protection measures. This however instills a new challenge in data-driven industries to look for new scientific solutions that can empower big discoveries while respecting the constraints of data privacy and governmental regulation.

An emerging solution is to leverage synthetic data~\cite{cramergan}, which statistically resembles real data and can comply with GDPR due to its synthetic nature. Generative Adversarial Network (GAN)~\cite{gan} is one of the emerging data synthesizing methodologies. 
Beyond its success in generating images~\cite{proven2021comicgan}, 
\cite{ctgan, tablegan, cramergan, ctabgan} have recently applied GAN to generate tabular data.
However, recent studies have shown that GANs may fall prey to membership inference attacks which greatly endanger the personal information present in the real training data~\cite{gan_leak,priv_mirage}. Therefore, it is imperative to safeguard the training of tabular GANs such that synthetic data can be generated without causing harm. To address these issues, prior work~\cite{pategan,long2019scalable,torkzadehmahani2019dp,torfi2020differentially} relies on differential privacy (DP)~\cite{dwork2008differential}. DP is a mathematical framework that provides theoretical guarantees bounding the statistical difference between any resulting ML model trained with or without a particular individual's information in the original training dataset. Typically, this can be achieved by injecting calibrated statistical noise while updating the parameters of a network during back-propagation, i.e., DP Stochastic Gradient Descent (DP-SGD)~\cite{abadi2016deep,xie2018differentially,chen2020gs}, or by injecting noise while aggregating teacher ensembles using the PATE framework~\cite{papernot2016semi,pategan}.
 
However, state-of-the-art (SOTA) tabular GAN algorithms only focus on two types of variables, namely continuous and categorical, overlooking an important class of \mixed data type. In addition, it is unclear if existing solutions can efficiently handle highly \imbalanced 
or skewed variables. Furthermore, most SOTA DP GANs are evaluated on images and their efficacy on tabular datasets needs to be verified. Existing DP GANs do not provide a well-defined consensus on which DP framework (i.e., DP-SGD or PATE) is optimal for training tabular GANs. Moreover, DP GAN algorithms such as \cite{chen2020gs} (GS-WGAN), \cite{pategan} (PATE-GAN) change the original GAN structure from one discriminator to multiple discriminators, which increases the complexity of the algorithm. And \cite{xie2018differentially}~(DP-WGAN) and \cite{torfi2020differentially}(RDP-GAN) use the weight clipping to bound gradients which introduces instability for GAN training.
 
 


In this paper, we extend \algo~\cite{ctabgan} to a new algorithm \algoplus. The objectives of \algoplus are two-folds: (1) further improve the synthetic data quality in terms of machine learning utility and statistical similarity; and (2) implement efficient DP into tabular GAN training to control its performance under different privacy budgets. To achieve the first goal, \algoplus introduces a new feature encoder used with \variables following single Gaussian distribution. Moreover, \algoplus adopts the Wasserstein distance plus gradient penalty (hereinafter referred to as \wgp) loss~\cite{wgan_gp} 
to further enhance the stability and effectiveness of GAN training. Finally, \algoplus adds a new auxiliary component to improve the synthesis performance for regression tasks.
To achieve the second goal, \algoplus uses DP-SGD algorithm to train a single instead of multiple discriminators as in PATE-GAN and GS-WGAN. This reduces the complexity of the algorithm.
Additionally, \algoplus reduces the privacy cost by accounting for sub-sampling~\cite{wang2019subsampled} of smaller subsets from the full dataset used to train the discriminator.

\begin{figure*}[t]
	\begin{center}
		\subfloat[\textit{bmi} in Insurance]{
			\includegraphics[width=0.25\textwidth]{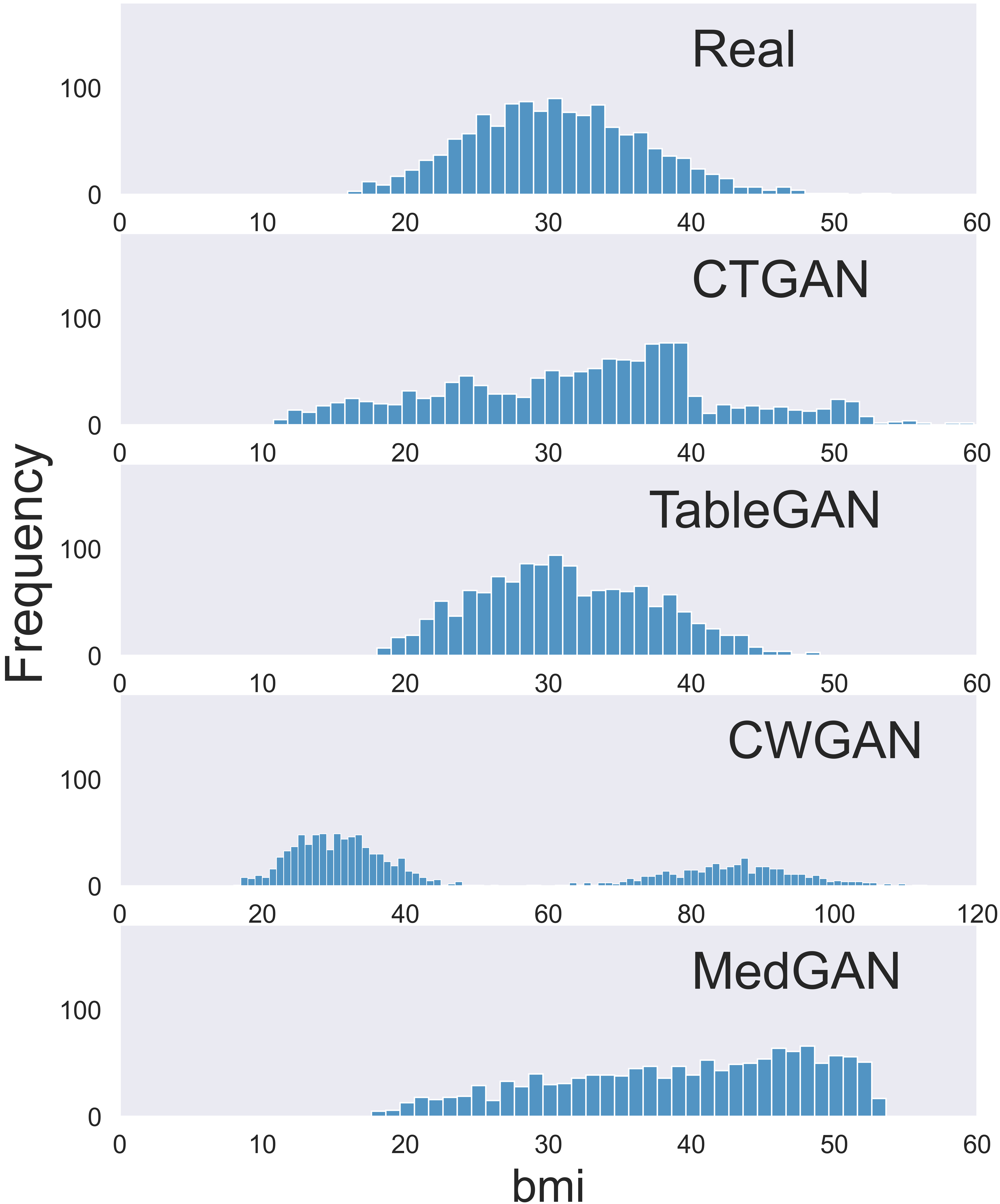}
			\label{fig:bmi_column_motivation}
		}
		\subfloat[\textit{Mortgage} in Loan]{
			\includegraphics[width=0.25\textwidth]{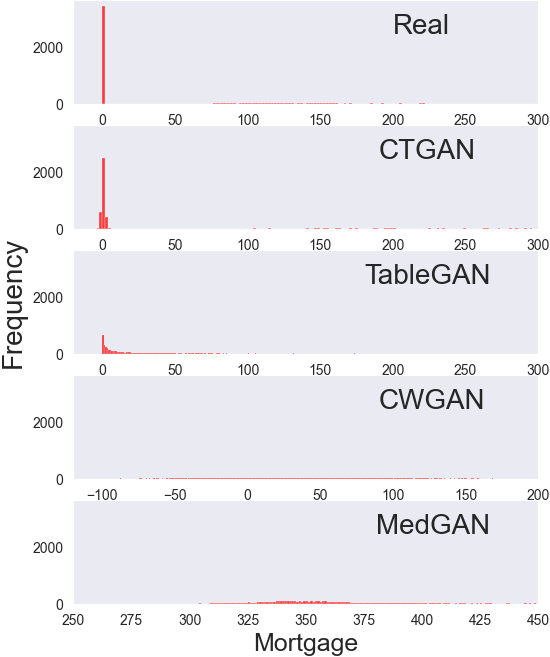}
			\label{fig:mortgage_column_motivation}
		}
		\subfloat[\textit{Amount} in Credit]{
			\includegraphics[width=0.22\textwidth]{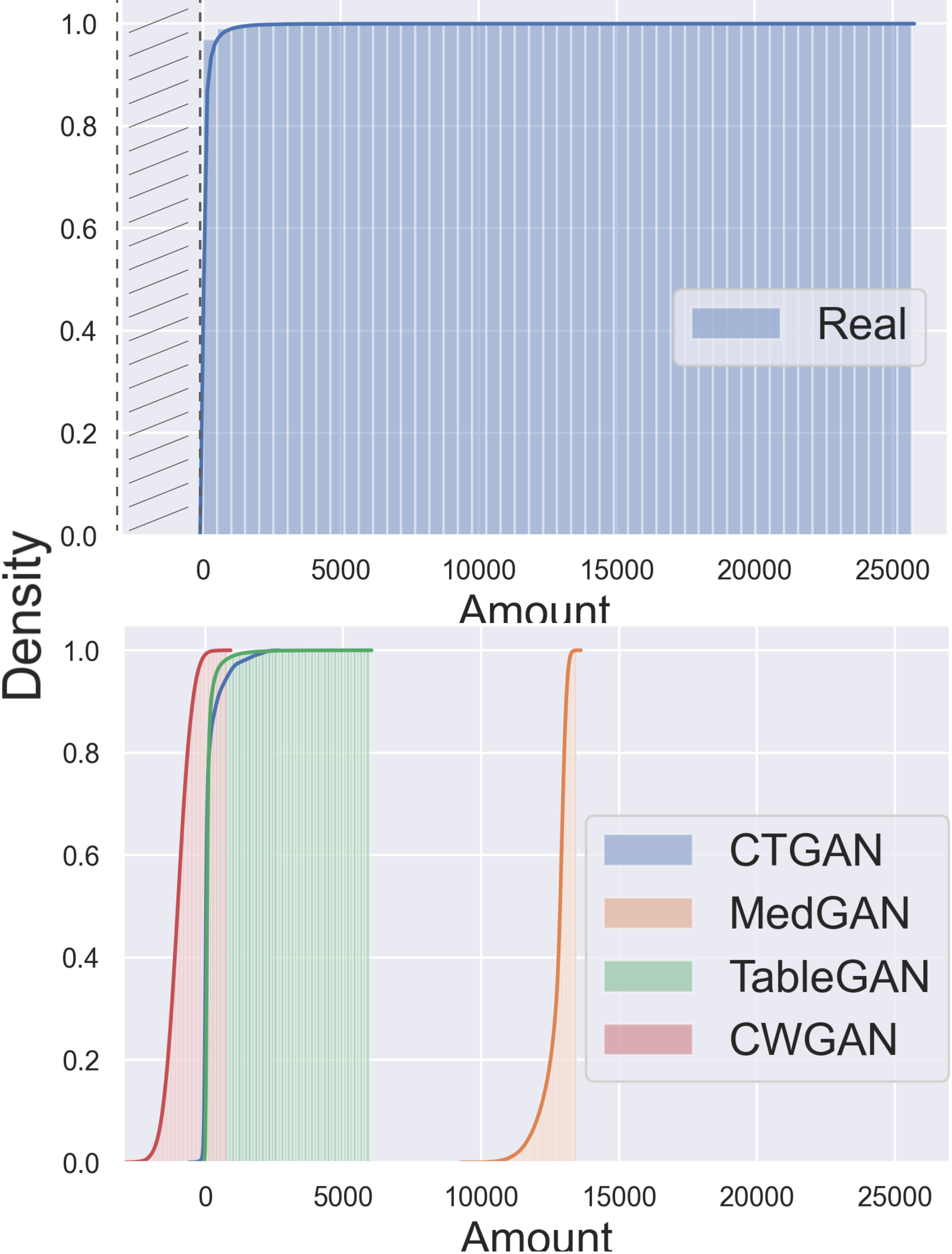}
			\label{fig:amount_result_motivation}
		}
		\subfloat[\textit{Hours-per-week} in Adult]{
			\includegraphics[width=0.25\textwidth]{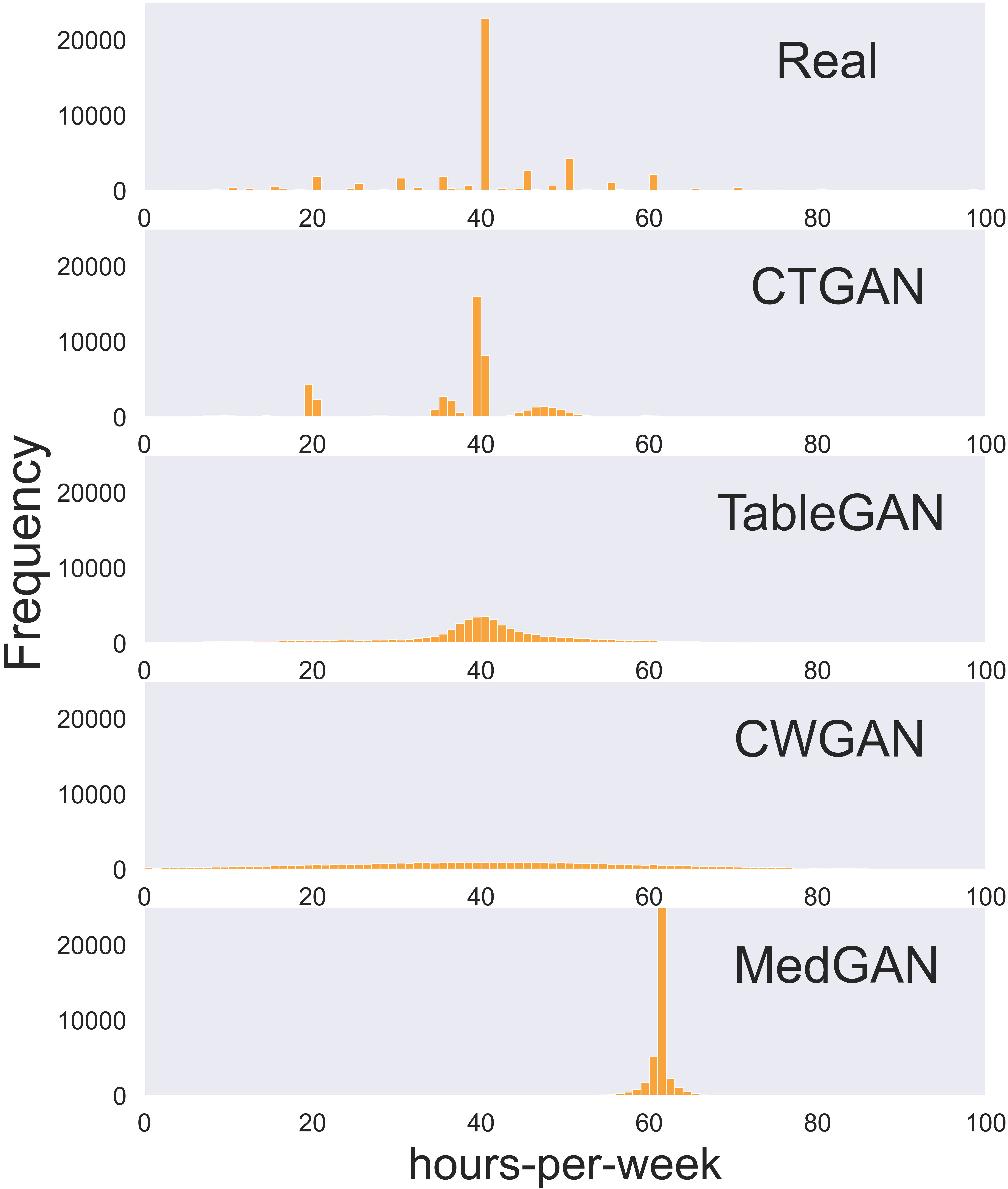}
			\label{fig:gmm_result_motivation}
		}
		\caption{ Challenges of modeling industrial dataset using existing GAN-based table generator: (a) single Gaussian (b) \bimodal type, (c) long tail distribution, and (d) \skewed data} 
		\label{fig:motivationcases}
 	\end{center}
\end{figure*}

We rigorously evaluate \algoplus using two setups: (1) without DP to generate data as realistic as possible; and (2) with DP under different privacy budgets to show the trade-off with data realism. Both setups rely on machine learning utility and statistical similarity of the synthetic data as evaluation metrics.
Specifically, \algoplus is tested on 7 widely used machine learning datasets: Adult, Covertype, Credit, Intrusion, Loan, Insurance and King against 7 SOTA GAN-based tabular data generation algorithms:  \ctgan, \tablegan, \cwgan and \medgan used in setup one, and PATE-GAN, DP-WGAN and GS-WGAN used in setup two. 
In setup one, \algoplus outperforms all baselines on average by at least 41.2\% on accuracy and 56.4\% on AUC. 
in setup two under the same privacy budget (i.e., $\epsilon = 1$ and $\epsilon = 100$ ), \algoplus outperforms all SOTA DP GANs on average by at least 48.16\% on accuracy and 38.05\% on F1-Score for the classification problem.

 
The main contributions of this study can be summarized as follows: (1) Novel conditional adversarial network which introduces a classifier/regressor providing additional supervision to improve the utility for ML applications. (2) Efficient modelling of continuous, categorical, and \textbf{\bimodal} variables via novel data encoding. 
(3) Improved GAN training using well-designed information, \dsm,  generator losses along with \wgp to enhance stability and effectiveness.
(4) Constructed a simpler and more stable DP GAN algorithm for tabular data to control its performance under different privacy budgets.

%% file: motivation_new.tex
\subsection{Motivation}
\label{sec:motivation}

We empirically demonstrate how the prior SOTA methods fall short in solving challenges in industrial data sets. The detailed experimental setup can be found in Sec.~\ref{ssec:setup}.

\textbf{Single Gaussian \columns}. Single mode Gaussian distributions are very common. Fig.~\ref{fig:motivationcases}(a) shows the histogram of \column \textit{bmi} (i.e., body mass index) in the Insurance dataset and synthetic data generated by 4 SOTA algorithms for this \column. The distribution of real data is close to a single mode Gaussian distribution. But except \tablegan, none of the SOTA algorithms can recover this distribution in their synthetic data. \ctgan uses variational Gaussian mixture (VGM) to model all continuous \columns. However, VGM is a complicated method to deal with single mode Gaussian distributions as it initially approximates the distribution with multiple Gaussian mixtures by default. \cwgan and \medgan use min-max normalization to scale the original data to [0, 1]. \tablegan also uses min-max normalization but scales the original data to [-1, 1] to better match the output of the generator using \textit{tanh} as activation function. The reason that min-max normalization works for \tablegan but not \medgan and \cwgan is because the training convergence for both algorithms is less stable than for \tablegan. However, since \tablegan applies min-max normalization on all \columns, it suffers from a disadvantage modelling column with complex multi-modal Gaussian distributions.

\textbf{\Mixed data type \columns}. To the best of our knowledge, existing GAN-based tabular generators only consider table columns as either categorical or continuous. However, in reality, a \column can be a mix of these two types, and often \columns have missing values. 
The \textit{Mortgage} \column from the Loan dataset is a good example of \bimodal variable.
Fig.~\ref{fig:motivationcases}(b) shows the distribution of the original and synthetic data generated by 4 SOTA algorithms for this \column.
According to the data description, a loan holder can either have no mortgage (0 value) or a mortgage (any positive value). In appearance, this \column is not a categorical type due to the numeric nature of the data. So all 4 SOTA algorithms treat this \column as continuous type without capturing the special meaning of the value zero. Hence, all 4 algorithms generate a value around 0 instead of exact 0. And the negative values for Mortgage have no/wrong meaning in the real world.

\textbf{Long tail distributions}. Many real world data can have long tail distributions where most of the occurrences happen near the initial value of the distribution, and rare cases towards the end.
Fig.~\ref{fig:motivationcases}(c) plots the cumulative frequency for the original (top) and synthetic (bottom) data generated by 4 SOTA algorithms for the
\textit{Amount} in the Credit dataset. This \column represents the transaction amount when using credit cards. One can imagine that most transactions have small amounts, ranging from a few bucks to thousands of dollars. However, there definitely exists a very small number of transactions with large amounts. Note that for ease of comparison both plots use the same x-axis, but Real has no negative values. 
Real data clearly has 99\% of occurrences happening at the start of the range, but the distribution extends until around $25000$. In comparison none of the synthetic data generators is able to learn and imitate this behavior.
    

\textbf{\Skewed multi-mode continuous variables}.
The term \textit{multi-mode} is extended from Variational Gaussian Mixtures (VGM). More details are given in Sec.~\ref{ssec:data_representation}. The intuition behind using multiple modes can be easily captured from Fig.~\ref{fig:motivationcases}(d). The figure plots in each row the distribution of the working \textit{Hours-per-week} \column from the Adult dataset.
This is not a typical Gaussian distribution.
There is an obvious peak at 40 hours but with several other lower peaks, e.g. at 50, 20 and 45. 
Also the number of people working 20 hours per week is higher than those working 10 or 30 hours per week.
This behavior is difficult to capture for the SOTA data generators (see subsequent rows in Fig.\ref{fig:motivationcases}(d)). The closest results are obtained by \ctgan which uses Gaussian mixture estimation for continuous variables. However, \ctgan loses some modes compared to the original distribution.
    

The above examples show the shortcomings of current SOTA GAN-based tabular data generation algorithms 
and motivate the design of our proposed \algoplus.


%% file: background_new.tex
\section{Related Work}
We divide the related work using GAN to generate tabular data into three: (i) based on GAN, (ii) based on conditional GAN, and (iii) based on DP GAN.

\textbf{GAN-based generator}. Several studies extend GAN to accommodate categorical variables by augmenting GAN architecture. 
MedGAN~\cite{medgan} combines an auto-encoder with a GAN. It can generate continuous or discrete variables, and has been applied to generate synthetic electronic health record (EHR) data. CrGAN-Cnet~\cite{cramergan} uses GAN to conduct Airline Passenger Name Record Generation. It integrates the Cramér Distance~\cite{cramerdistance} and Cross-Net architecture~\cite{crossnet} into the algorithm. In addition to generating with continuous and categorical data types,  CrGAN-Cnet can also handle missing value in the table by adding new \columns. TableGAN~\cite{tablegan} introduces information loss and a classifier into GAN framework. It specifically adopts Convolutional Neural Network (CNN) for generator, discriminator and classifier. 
Although aforementioned algorithms can generate tabular data, they cannot specify how to generate from a specific class for particular variable. For example, it is not possible to generate health record for users whose sex is female. 


\textbf{Conditional GAN-based generator}. Due to the limitation of controlling generated data via GAN, Conditional GAN is increasingly used, and its conditional vector can be used to specify to generate a particular class of data. This feature is important when our available data is limited and highly skewed, and we need synthetic data of a specific class to re-balance the distribution. For instance, for preparing starting dataset of online learning scenarios~\cite{Zhao:DSN19,Zhao:TDSC21,younesian2020qactor}.
CW-GAN~\cite{cwgan} applies the Wasserstein distance~\cite{wgan} into the conditional GAN framework. It leverages the usage of conditional vector to oversample the minority class to address imbalanced tabular data generation. CTGAN~\cite{ctgan} integrates PacGAN~\cite{pacgan} structure in its discriminator and uses Generator loss and WGAN loss plus gradient penalty (GP)~\cite{wgan_gp} to train a conditional GAN framework. It also adopts a strategy called training-by-sampling, which takes advantage of conditional vector, to deal with the imbalanced categorical variable problem. 

\algoplus not only focuses on modelling both continuous and categorical variables, but also covers the \mixed data type (i.e., \columns that contain both categorical and continuous values, or even missing values). We effectively combine the strengths of the prior art, such as \wgp, classifier, information and generator losses along with an effective encoding. Furthermore, we proactively address the pain points of single Gaussian and long tail \variable distributions 
and propose a new conditional vector structure to better deal with imbalanced datasets.

\textbf{Differential Private Tabular GANs}.
To avoid leaking sensitive information on single individuals, previous studies explore multiple differential private learning techniques applied to GANs.
Table~\ref{table:DPGANS} provides an overview. 
PATE-GAN~\cite{pategan} uses PATE~\cite{papernot2016semi} which relies on output sanitization by perturbing the output of an ensemble of teacher discriminators via Laplacian noise to train a student discriminator scoring the generated samples. 
One key limitation is that the student discriminator only sees synthetic data. Since this data is potentially unrealistic, the provided feedback can be unreliable. \cite{xie2018differentially}~(DP-WGAN), \cite{chen2020gs}~(GS-WGAN) and \cite{torfi2020differentially}(RDP-GAN) use differential private stochastic gradient descent (DP-SGD) coupled with the Wasserstein loss. Moreover, DP-WGAN uses a momentum accountant whereas GS-WGAN and RDP-GAN use a R\'enyi Differential Privacy (RDP) accountant. 
The Wasserstein loss is known to be more effective against mode-collapse compared to KL divergence~\cite{pmlr-v70-arjovsky17a}. The RDP accountant provides tighter bounds on the privacy costs improving the privacy-utility trade-off.
To incorporate differential privacy guarantees and make the training compatible with the Wasserstein Loss, \cite{xie2018differentially,torfi2020differentially} use weight clipping to enforce the Lipschitz constraint. The drawback is the need for careful tuning of the clipping parameter (see Sec.~\ref{ssec:dp-sgd}). To overcome this issue, \cite{chen2020gs} enforces the Lipschitz constraint via a gradient penalty term as suggested by~\cite{gulrajani2017improved}, but addresses only images which are a better fit for GANs and studies its efficacy only for training the generator network. 

The proposed \algoplus leverages RDP-based privacy accounting comparing to PATE used by PATE-GAN. 
Same as DP-WGAN, \algoplus uses one discriminator instead of multiple ones trained by PATE-GAN and GS-WGAN. Since \algoplus adopts \wgp loss, it intrinsically constraints the gradient norm allowing to forgo the weight clipping used in DP-WGAN. This leads to a more stable training. In a nutshell, \algoplus by training only one discriminator with \wgp loss results in a more stable DP GAN algorithm compared to the SOTA algorithms.



\color{black}
\begin{table}[t]
\caption{Overview of DP GANs}
\resizebox{\columnwidth}{!}{
\centering
\begin{tabular}{|c|c|c|c|c|c|c|c|c|}
\hline
\textbf{Model} & \textbf{DP Algo} & \textbf{Loss} & \textbf{WC} &\textbf{DP Site} & \textbf{Single $\mathcal{D}$ } & \textbf{Noise}& \textbf{Account.} & \textbf{Data Format}\\ 
\hline
PATE-GAN & PATE & KL Diver. & No &  $\mathcal{D}$ & No  & $\text{Lap.}$  & PATE & Table\\\hline
DP-WGAN & DP-SGD & Was.   & Yes   &  $\mathcal{D}$ & Yes   & $\mathcal{N}$  & Moment & Image $\&$ Table \\ \hline
GS-WGAN & DP-SGD & Was. + GP  & No & $\mathcal{G}$ & No  & $\mathcal{N}$ & RDP & Image\\ \hline

RDP-GAN & DP-SGD & Was. & Yes & $\mathcal{D}$  & Yes & $\mathcal{N}$ & RDP & Table\\ \hline
\algoplus & DP-SGD & Was. + GP  & No & $\mathcal{D}$ & Yes & $\mathcal{N}$ & RDP & Table\\ \hline
\multicolumn{9}{l}{$^*$ Was. (Wasserstein), WC (Weight Clipping), GP (Gradient Penalty), Lap. (Laplacian noise)} \\
\multicolumn{9}{l}{$^*$ $\mathcal{N}$ (Gaussian noise), $\mathcal{G}$ (Generator), $\mathcal{D}$ (Discriminator)} \\
\end{tabular}
}
\label{table:DPGANS}
\end{table}

%% file: model_v2_new.tex
\section{\algoplus}

\begin{figure*}[t]
    \centering
    \includegraphics[width=0.8\linewidth]{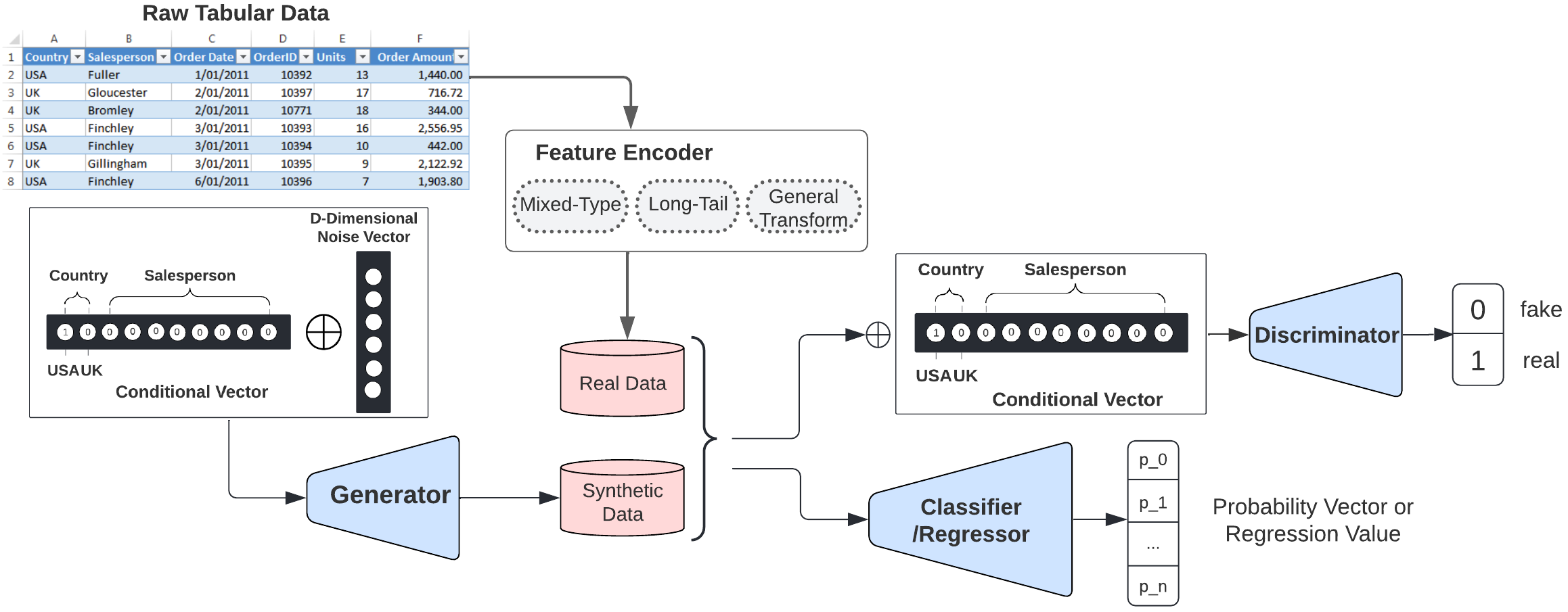}
    \caption{Synthetic Tabular Data Generation via \algoplus}
    \label{fig:STD}
\end{figure*}

\algoplus is a tabular data generator designed to overcome the challenges outlined in Sec.~\ref{sec:motivation}. \algoplus adopts a re-designed min-max scaler to normalize single Gaussian \column. We also propose a novel \textit{\encoder} which can better represent mixed categorical-continuous variables as well as missing values. \algoplus is based on a conditional GAN (CGAN) to efficiently treat minority classes, with the addition of \wgp, \dsm, information and generator losses (\cite{wgan_gp,tablegan,acgan,ctgan}) to improve data quality and training stability. Moreover, we leverage a log-frequency sampler to overcome the mode collapse problem for imbalanced \columns. Finally, differential private SGD training is implemented for the discriminator to achieve strict privacy guarantees.

\ifshowntationtable
Before illustrating \algoplus, we summarized the used notation in Table~\ref{tab:symboltable}.

\begin{table}
\centering
\caption{Notation}
\label{tab:symboltable}
\begin{tabular}{|C{1.5cm} |L{6cm}|} \hline
\textbf{Symbol}	& \textbf{Description}\\ \hline
$\mathcal{G}$	& Generator of \algoplus					 \\\hline
$\mathcal{D}$ 	& Discriminator of \algoplus					 \\\hline
$\mathcal{C}$	& Classifier of \algoplus					 \\\hline
$\mathcal{V}$	& Conditional vector 
\\\hline
$\mathcal{N}$ 	& Normal distribution					 \\\hline
$\mu_i$ 	& value of categorical mode, mean value of Gaussian mixture mode					 \\\hline
$\rho_i$ 	& probability density calculated based on mode $i$					 \\\hline
$\gamma_{i,j}$&  one-hot encoding vector for $i$th row value in categorical \column $j$			\\\hline
$\bigoplus$&  vector concatenation			\\\hline
\end{tabular}
\end{table}
\fi

\subsection{Technical Background}

\subsubsection{Tabular GAN}
GANs are a popular method to generate synthetic data first applied with great success to images~\cite{stylegan} and later adapted to tabular data~\cite{yahi_gan}. GANs leverage an adversarial game between a generator trying to synthesize realistic data and a discriminator trying to discern synthetic from real samples.

To address the problem of dataset imbalance, we leverage \textit{conditional generator} and \textit{training-by-sampling} methods from CTGAN. The idea behind this is to use an additional vector, termed  conditional vector, to represent the classes of categorical \columns. This vector is both fed to the generator and used to bound the sampling of the real training data to subsets satisfying the condition. We can leverage the condition to resample all classes giving higher chances to minority classes to train the model. 
To improve the stability of GAN training, \algoplus adopts \wgp~\cite{wgan_gp} loss. Previous study WGAN~\cite{wgan} offers stability in training GAN. However, the use of gradient clipping leads to issues such as exploding and vanishing gradients. Comparing to WGAN~\cite{wgan}, WGAN-GP replaces weight clipping with a constraint on the gradient norm of the discriminator to enforce Lipschitz continuity. This further stabilizes the training of the network and requires less hyper-parameter tuning. Another unique feature of WGAN-GP is that its discriminator updates 5 times per mini-batch data comparing to only 1 time of generator. This influences our differential privacy budget (see details in Sec.~\ref{ssec:dp-sgd}).
To enhance the generation quality, we incorporate three extra terms into the loss function of the generator: information~\cite{tablegan}, \dsm  (referred as classification loss in \cite{acgan} for classification problems) and generator loss~\cite{ctgan}. 
The information loss penalizes the discrepancy between statistics of the generated data and the real data. This helps to generate data which is statistically closer to the real one.
The \dsm loss requires adding to the GAN architecture an auxiliary classifier (or regressor) in parallel to the discriminator. 
For each synthesized value the classifier (or regressor) outputs a predicted value.
The \dsm loss quantifies the discrepancy between the synthesized and predicted values in the downstream analysis. This helps  increase the semantic integrity of synthetic records.
For instance, for a classification dataset, (sex=female, disease=prostate cancer) is not a semantically correct record as women do not have a prostate, and no such record should appear in the original data and is hence not learnt by the classifier.
The generator loss measures the difference between the given conditions and the output classes of the generator. This loss helps the generator to learn to produce the exact same classes as the given conditions. \Dsm loss is used by TableGAN but not by CTGAN, since CTGAN does not contain a classifier. Whereas, the generator loss is implemented by CTGAN but not by TableGAN, as TableGAN is not a conditional GAN. Both only treat  classification problems.

To counter complex distributions in continuous \columns we embrace the \textit{Mode-Specific Normalization (MSN)} idea~\cite{ctgan} which encodes each value as a value-mode pair stemming from the Gaussian mixture model.



\subsubsection{Differential Privacy}




DP is becoming the standard solution for privacy protection and has even been adopted by the US census department to bolster privacy of citizens~\cite{Hawes2020Implementing}. DP protects against privacy attacks by  minimizing the influence of any individual data point based on a given privacy budget. In this work, we leverage the R\'enyi Differential Privacy (RDP)~\cite{mironov2017renyi} as it provides stricter bounds on the privacy budget. A randomized mechanism \( \mathcal{M} \) is $(\lambda,\epsilon)$-RDP with order $\lambda$, if {\small
$D_{\lambda}(\mathcal{M}(S)||\mathcal{M}(S')) = \frac{1}{\lambda-1}log\mathbb{E}_{x\sim\mathcal{M}(S)} \left[  \left(\frac{P[\mathcal{M}(S)=x]}{P[\mathcal{M}(S')=x]} \right) \right]^{\lambda-1}\leq\epsilon$
holds for any adjacent datasets $S$ and $S'$, where {\small
$D_\lambda(P||Q)=\frac{1}{\lambda-1}log\mathbb{E}_{x\sim Q}[(P(x)/Q(x))^\lambda]$}} represents the R\'enyi divergence. In addition, a $(\lambda,\epsilon)$-RDP mechanism \( \mathcal{M} \) can be expressed as:
\begin{equation}
    \label{eq:rdp-to-dp}
    (\epsilon+\frac{log1/\delta}{\lambda-1},\delta)\text{-DP}.
\end{equation}
For the purpose of this work $\mathcal{M}$ corresponds to a tabular GAN model with privacy budget $(\lambda, \epsilon)$.

RDP is a strictly stronger privacy definition than DP as it provides tighter bounds for tracking the cumulative privacy loss over a sequence of mechanisms via the Composition theorem~\cite{mironov2017renyi}. Let $\circ$ denote the composition operator. For \( \mathcal{M}_{1} \),...,\( \mathcal{M}_{k} \) all being $(\lambda,\epsilon_i)$-RDP, the composition \( \mathcal{M}_{1} \)$\circ ... \circ$\( \mathcal{M}_{k} \) is
\begin{equation}
    \label{eq:composition}
    (\lambda,\sum_{i}\epsilon_{i})\text{-RDP}
\end{equation}

Additionally, for a Gaussian Mechanism\cite{dwork2014algorithmic} \( \mathcal{M}_{\sigma} \) parameterized by $\sigma$ as:
\begin{equation}
    \label{eq:gaussian_mechanism}
    \mathcal{M}_{\sigma}(x) = f(x) + \mathcal{N}(0,\sigma^{2}I)
\end{equation}
where $f$ denotes an arbitrary function with sensitivity $\Delta_{2f} = \max_{S,S'}||f(S) - f(S')||_{2}$ over all adjacent datasets $S$ and $S'$, and  $\mathcal{N}$ represents a Gaussian distribution with zero mean and covariance $\sigma^{2}I$ (where $I$ is the identity matrix),
$\mathcal{M}_{\sigma}$ satisfies $(\lambda,\frac{\lambda\Delta_{2f}^{2}}{2\sigma^{2}})$-RDP~\cite{mironov2017renyi}. 

Lastly, two more theorems are key to this work. The post processing theorem~\cite{dwork2014algorithmic} states that if \( \mathcal{M} \) satisfies $(\epsilon,\delta)$-DP, $F\circ\mathcal{M}$ will satisfy $(\epsilon,\delta)\text{-DP}$, where $F$ can be any arbitrary randomized function. Hence, it suffices to train one of the two networks in the GAN architecture with DP guarantees to ensure that the overall GAN is compatible with differential privacy. RDP for subsampled mechanisms~\cite{wang2019subsampled} computes the reduction in privacy budget when sub-sampling private data. Formally, let $\mathcal{X}$ be a dataset with $n$ data points and \textbf{subsample} return $m \leq n$ subsamples without replacement from $\mathcal{X}$ (subsampling rate $\gamma = m/n$).
For all integers $\lambda \geq 2 $, if a randomized mechanism $\mathcal{M}$ is $(\lambda,\epsilon(\lambda))$-RDP, then $\mathcal{M}\circ\textbf{subsample}$ is
\begin{equation}
    \label{eq:subsampling}
    (\lambda,\epsilon'(\lambda))\text{-RDP}
\end{equation}

where {\small $\epsilon'(\lambda) \leq \frac{1}{\lambda-1}log(1 + \gamma^{2}\binom{\lambda}{2}\min\left\{4(e^{\epsilon(2)}-1),e^{\epsilon(2)}\min\{2,(e^{\epsilon(\infty)}-1)^{2}\}\right\} \allowbreak$ + \\$\sum_{j=3}^{\lambda}\gamma^{j} \binom{\lambda}{j}e^{(j-1)\epsilon(j)}\min\{2,(e^{\epsilon(\infty)-1)^{j})}\})$}
\color{black}

\subsection{Architecture of \algoplus}
\label{sec:design_algorithm}
The structure of \algoplus is shown in Fig.~\ref{fig:STD}. It comprises three blocks: Generator $\mathcal{G}$, Discriminator $\mathcal{D}$ and an auxiliary component (either a classifier or a regressor) $\mathcal{C}$. 
Since our algorithm is based on conditional GAN, the generator requires a noise vector plus a conditional vector. Details on the conditional vector are given in Sec.~\ref{sec:condvec}. Before feeding data to $\mathcal{D}$ and $\mathcal{C}$, \variables are encoded via different feature encoders depending on the \variable type and characteristics. Details of the used encoders are provided in Sec.~\ref{ssec:data_representation}, \ref{ssec:gt} and \ref{ssec:longgtail}.

GANs are trained via a zero-sum min-max game where the discriminator tries to maximize the objective, while the generator tries to minimize it. The game can be seen as a mentor ($\mathcal{D}$) providing feedback to a student ($\mathcal{G}$) on the quality of his work. 
Here, we introduce additional feedback for $\mathcal{G}$ based on the information loss, \dsm loss and generator loss. The information loss matches the first-order (i.e., mean) and second-order (i.e., standard deviation) statistics of synthesized and real records. This leads the synthetic records to have the same statistical characteristics as the real records.
The \dsm loss equates the correlation between target \column and the other \column values. 
This helps to check the semantic integrity, and penalizes synthesized records where the combination of values are semantically incorrect.
Finally, the generator loss is the cross-entropy between the given conditional vector and the generated output classes. It enforces the conditional generator to produce the same classes as the given conditional vector. 
These three losses add to the default loss term (i.e., \wgp) of $\mathcal{G}$ during training.
$\mathcal{G}$ and $\mathcal{D}$ are implemented using CNNs with the same structure as in~\cite{tablegan}.
CNNs are good at capturing the relation between pixels within an image, which in our case, can help to increase the semantic integrity of synthetic data. To process row records stored as vectors with CNN, we wrap the row data into the closest square matrix dimensions, i.e. $d \times d$ where $d$ is the ceiled square root of the row data dimensionality and pad missing values with zeros.
$\mathcal{C}$ uses a multi-layer-perceptron (MLP) with four 256-neuron hidden layers. The classifier is trained on the original data to better interpret the semantic integrity. Hence synthetic data are reverse transformed from their matrix encoding to vector (details in Sec.~\ref{ssec:data_representation}). Real data is encoded (details in Sec.~\ref{ssec:data_representation} and \ref{ssec:longgtail}) before being used as input for $\mathcal{C}$ to create the class label predictions.

Let $f_x$ and $f_{\mathcal{G}(z)}$ denote the features fed into the softmax layer of $\mathcal{D}$ for a real sample $x$ and a sample generated from latent value $z$, respectively. The \textbf{information loss} for $\mathcal{G}$ is expressed as
$\mathcal{L}_{info}^{G} = ||\mathbb{E}[f_x]_{x \sim p_{data}(x)} - \mathbb{E}[f_{\mathcal{G}(z)}]_{z \sim p(z)}||_{2} + ||\mathbb{SD}[f_x]_{x \sim p_{data}(x)} - \mathbb{SD}[f_{\mathcal{G}(z)}]_{z \sim p(z)}||_{2}$ where $p_{data}(x)$ and $p(z)$ denote prior distributions for real data and latent variable, $\mathbb{E}$ and $\mathbb{SD}$ denote the mean and standard deviations of the features, respectively. 
The \textbf{\dsm loss} is given by 
$\mathcal{L}_{dstream}^{G} = \mathbb{E}[|l(G(z))-\mathcal{C}(fe(G(z)))|]_{z \sim p(z)}$
where $l(.)$ returns the target \variable and $fe(.)$  returns the input features of a given row $x$.
Finally, the \textbf{generator loss} is given by  $\mathcal{L}_{generator}^{G} = H(m_{i}, \hat{m}_{i})$ where $m_{i}$ and $\hat{m}_i$ are the given and generated conditional vector bits corresponding to column $i$ and $H(.)$ is the cross-entropy loss. Columns are selected using the training-by-sampling procedure (see Sec.~\ref{sec:condvec} for details).


Let $\mathcal{L}^{D}_{default}$ and $\mathcal{L}^{G}_{default}$ denote the GAN loss of discriminator and generator from \wgp where its unique objective function of discriminator is defined as follows: 
\begin{align*}
    L = \underbrace{\underset{\tilde{x} \sim \mathbb{P}_g}{\mathbb{E}} [D(\tilde{x})] - \underset{x \sim \mathbb{P}_r}{\mathbb{E}} [D(x)]}_\text{original discriminator loss} + \underbrace{\lambda \underset{\hat{x} \sim \mathbb{P}_{\hat{x}}}{\mathbb{E}} [ (\left\lVert \nabla_{\hat{x}} D(\hat{x}) \right\rVert_{2} - 1])^2}_\text{gradient penalty}
\end{align*}
where $\mathbb{P}_{\hat{x}}$ is defined as sampling uniformly along straight lines between pairs of points sampled from the real data distribution $\mathbb{P}_r$ and the generator distribution $\mathbb{P}_g$.
For $\mathcal{G}$ the complete training objective is $\mathcal{L}^{G}=\mathcal{L}^{G}_{default}+\mathcal{L}_{info}^{G}+\mathcal{L}_{dstream}^{G} + \mathcal{L}_{generator}^{G}$. The training objective for $\mathcal{D}$ is unchanged. 
Finally, the loss to train the auxiliary $\mathcal{C}$ is similar to the \dsm loss of the generator, i.e. $\mathcal{L}_{dstream}^{C} = \mathbb{E}[|l(x)-\mathcal{C}(fe(x))|]_{x \sim p_{data}(x)}$.

\subsection{\encoder}
\label{ssec:data_representation}


\begin{figure}[t]
	\begin{center}
		\subfloat[Mixed type \column distribution with VGM]{
			\includegraphics[width=0.47\columnwidth]{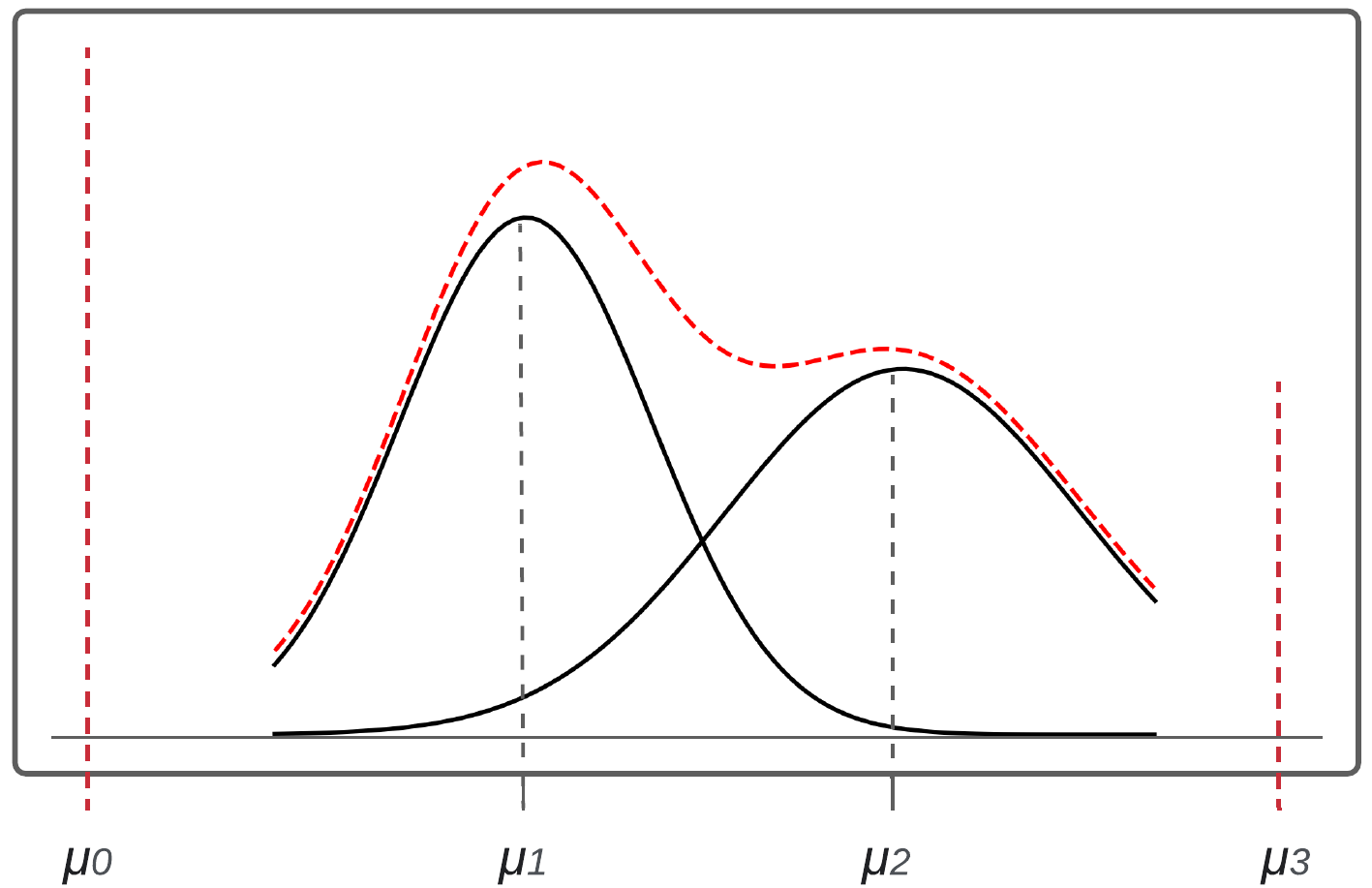}
			\label{fig:vgm}
		}
		\hfil
		\subfloat[Mode selection of single value in continuous \column]{
			\includegraphics[width=0.47\columnwidth]{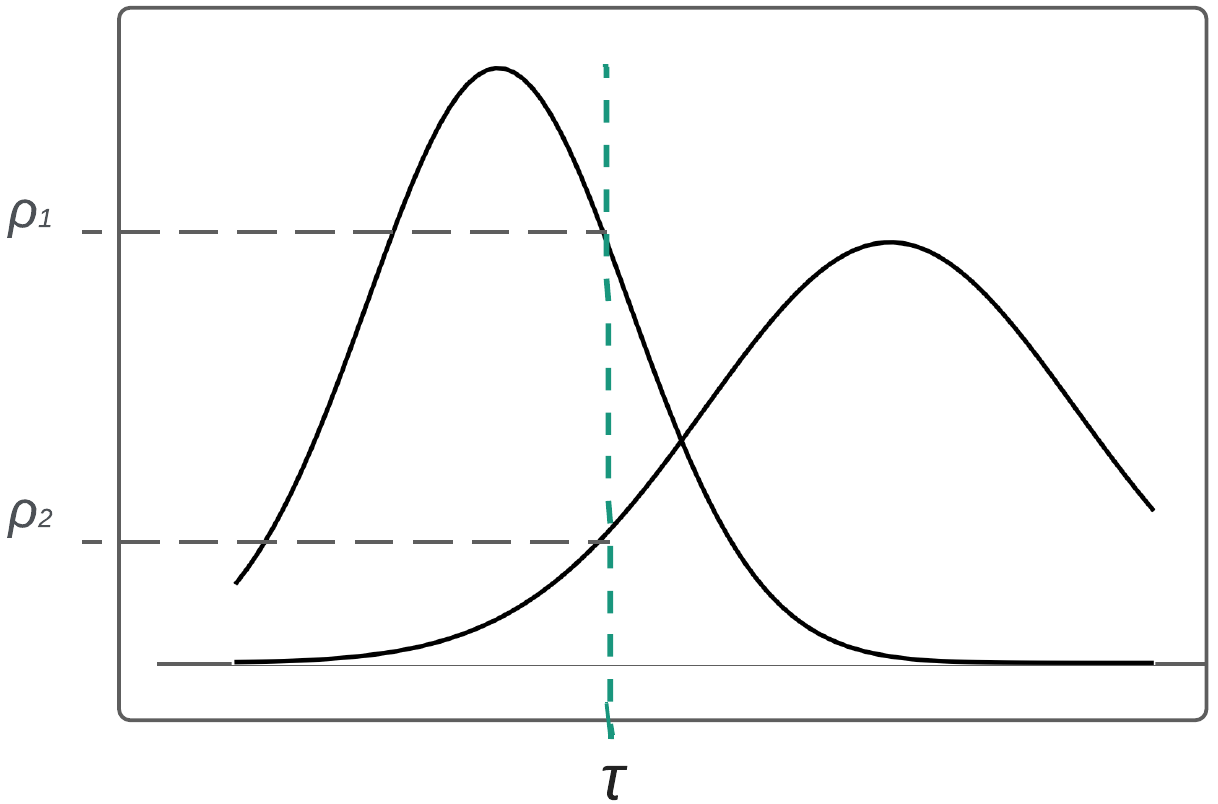}
			\label{fig:vgm_single}
		}
		\caption{Encoding for mix data type \column}
		\label{fig:gmm_distribution}
	\end{center}
\end{figure}


The tabular data is encoded \column by \column. We distinguish three types of \columns: categorical, continuous and \mixed. 
We define \columns as mixed if they contain both categorical and continuous values or continuous values with missing values. We propose the new \encoder to deal with such \columns. With this encoder, values of \mixed \columns are seen as concatenated value-mode pairs. We illustrate the encoding via the exemplary distribution of a \mixed \column shown in red in Fig.~\ref{fig:gmm_distribution}(a). One can see that values can either be exactly $\mu_0$ or $\mu_3$ (the categorical part) or distributed around two peaks in $\mu_1$ and $\mu_2$ (the continuous part). 
We treat the continuous part by adapting the \textit{Mode-Specific Normalization} (MSN) idea from~\cite{ctgan} in using a variational Gaussian mixture model (VGM)~\cite{prml} to estimate the number of modes $k$, e.g. $k=2$ in our example, and fit a Gaussian mixture. The learned Gaussian mixture is $\mathbb{P} = \sum_{k=1}^{2} \omega_k \mathcal{N}(\mu_k, \sigma_k)$, 
where $\mathcal{N}$ is the normal distribution and $\omega_k$, $\mu_k$ and $\sigma_k$ are the weight, mean and standard deviation of each mode, respectively.

To encode values in the continuous region of the \column distribution, we associate and normalize each value with the mode having the highest probability (see Fig.~\ref{fig:gmm_distribution}(b)). Given $\rho_1$ and $\rho_2$ being the probability density from the two modes in correspondence of the \column value $\tau$ to encode, we select the mode with the highest probability. In our example $\rho_1$ is higher and we use mode $1$ to normalize $\tau$. The normalized value $\alpha$ is: $\alpha =  \frac{\tau - \mu_1}{4\sigma_1}$.
Moreover we keep track of the mode $\beta$ used to encode $\tau$ via one-hot encoding, e.g.  $\beta = [0,1,0,0]$ in our example. The final encoding is giving by the concatenation of $\alpha$ and $\beta$: $\alpha \bigoplus \beta$
\ifshowntationtable
.
\else
where $\bigoplus$ is the vector concatenation operator.
\fi
The categorical part (e.g., $\mu_0$ or $\mu_3$ in Fig.~\ref{fig:gmm_distribution}(a)) is treated similarly, except $\alpha$ is directly set to 0. Because the category is determined only by one-hot encoding part. For example, for a value in $\mu_3$, the final encoding is given by $0 \bigoplus [0, 0, 0, 1]$.
Categorical \columns use the same encoding as the continuous intervals of \mixed \columns. 
Categorical \columns are encoded via a one-hot vector $\gamma$. Missing values are treated as a separate unique class and we add an extra bit to the one-hot vector for it. A row with $[1, \dots, N]$ \columns is encoded by concatenation of the encoding of all \column values, i.e. either $(\alpha \bigoplus \beta)$ for continuous and mixed \columns or $\gamma$ for categorical \columns. 
Having $n$ continuous/mixed \columns and $m$ categorical \columns ($n + m = N$) the final encoding is: 
\begin{equation}
\label{condvec}
    \bigoplus_{i=1}^{n} \alpha_i\mathsmaller{\bigoplus} \beta_{i} \;
    \bigoplus_{j=n+1}^{N} \gamma_{j} 
\end{equation}

\subsection{Counter \Imbalanced Training Datasets}
\label{sec:condvec}

\begin{figure}[t]
    \centering
    \includegraphics[scale=.18]{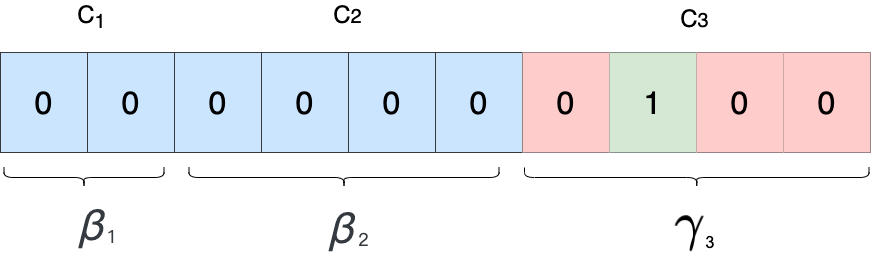}
    \caption{Conditional vector: example selects class 2 from third \column out of three}
    \label{fig:condvec}
\end{figure}

In \algoplus, we use conditional GAN to counter imbalanced training datasets using training-by-sampling~\cite{ctgan}, but extended to include the modes of continuous and mixed columns.
When we sample real data, we use the conditional vector to filter and rebalance the training data. 
The conditional vector $\mathcal{V}$ is a bit vector given by the concatenation of all mode one-hot encodings $\beta$ (for continuous and \mixed \columns) and all class one-hot encodings $\gamma$ (for categorical \columns) for all \columns present in Eq.~\eqref{condvec}. Each conditional vector specifies a single mode or a class. More in detail, $\mathcal{V}$ is a zero vector with a single one in correspondence to the selected \column with selected mode/class.
Fig.~\ref{fig:condvec} shows an example with three \columns, one continuous ($C_1$), one \mixed ($C_2$) and one categorical ($C_3$), with class 2 selected on $C_3$. 

To rebalance the dataset, each time we need a conditional vector during training, we first randomly choose a \column with uniform probability. Then we calculate the probability distribution of each mode (or class for categorical \columns) in that \column using frequency as proxy and sample a mode based on the logarithm of its probability. Using the log probability instead of the original frequency gives minority modes/classes higher chances to appear during training. This helps to alleviate the collapse issue for rare modes/classes. Extending the conditional vector to include the continuous and \mixed \columns helps to deal with imbalance in the frequency of modes used to represent them.
Moreover, since generator is conditioned on all data-types during training,  this enhances the learned correlation between all \columns.

\subsection{General Transform} 
\label{ssec:gt}

\algo originally adopts the mode-specific-normalization (MSN) from \ctgan to encode all continuous \variables. MSN uses VGM to estimate the distribution of continuous \variables. Fig.~\ref{fig:motivationcases}(a) shows that VGM is not suitable for simple distributions such as single Gaussian. Another problem is the dimensionality explosion caused by using one-hot-encoding for categorical variables with a high number of categories.
To counter both problems we propose the general transform (GT). GT is an effective approach to minimize the complexity of our algorithm. 


The main idea of GT is to encode columns in the range of $(-1,1)$. This makes the encoding directly compatible with the output range of the generator using \textit{tanh} activation function. This is achieved via a shifted and scaled min-max normalization.
Mathematically, 
given a data point $x_i$ of a continuous \column $x$, the transformed value, $x_i^t = 2*\frac{x_i - min(x)}{max(x)-min(x)} - 1$ where $min(x)$ and $max(x)$ represents the minimum and maximum values of the continuous \column. Inversely an encoded or generated value $x_i^t$ may be reverse transformed as $X_i=(max(x)-min(x))*\frac{X_i^t+1}{2}+min(x)$.
Continuous variable can be directly treated with the above formulas for normalization and denormalization. Categorical variables are first encoded using integers before using the above normalization and rounded to integers after using the above denormalization.


A similar transform was first introduced by \tablegan, but it applies this transformation on all \columns. This choice is not optimal. From our experiments, we find that this technique only works well for continuous columns with simple distributions such as a single-mode Gaussian and does not cater to more complex distributions. By default, \algoplus deals with continuous \variable with MSN and only selectively uses GT for processing single-mode Gaussian \columns. 
Similarly, categorical columns should prefer MSN as encoding rather than GT. Using GT loses the
mode indicator, i.e. $\beta_1$ in Fig.~\ref{fig:condvec}, from the conditional vector forgoing the ability to enhance the correlation between \columns for specific categories. Moreover, using integers instead of one-hot vectors can impose artificial distances between the different categories which do not reflect the reality. Therefore, we recommend to use GT for categorical \columns only if the categorical \variables contain so many categories that the available machines can not train with the encoded data.

\subsection{Treat Long Tails}
\label{ssec:longgtail}
We encode continuous values using variational Gaussian mixtures to treat multi-mode data distributions (details in Sec.~\ref{ssec:data_representation}).
However, Gaussian mixtures can not deal with all types of data distribution, notable distributions with long tail where few rare points are far from the bulk of the data.
VGM has difficulty to encode the values towards the tail.
To counter this issue we pre-process \columns with long tail distributions with a logarithm transformation. For such a \column having values with lower bound $l$, we replace each value $\tau$ with compressed $\tau^c$:
\begin{equation}
\label{eq:preprossesing}
\tau^c =  \left\{
	\begin{array}{rl}
		 \mbox{log($\tau$)} &  \mbox{if $l>$0} \\
		\mbox{log($\tau$ - $l$+$\epsilon$)} & \mbox{if $l\leqslant$0}  \mbox{, where } \mbox{$\epsilon>$0} 	 
	\end{array} \right\}
\end{equation}

The log-transform allows to compress and reduce the distance between the tail and bulk data making it easier for VGM to encode all values, including tail ones. We show the effectiveness of this simple yet performant method in Sec.~\ref{sec:further}.

\subsection{Differential Privacy}
\label{ssec:dp-sgd}
DP-SGD~\cite{abadi2016deep} is the central framework to provide DP guarantees in this work. DP-SGD uses noisy stochastic gradient descent to limit the influence of individual training samples $x_i$. After computing the gradient $g(x_i)$, the gradient is clipped based on a clipping parameter $C$ and its L2 norm  $\bar{g}(x_i) \leftarrow g(x_i)/\max(1,\frac{||g(x_i)||_{2}}{C})$, and Gaussian noise is added $\Tilde{g}(x_i)\leftarrow \bar{g}(x_i)+\mathcal{N}(0,\sigma^{2}C^{2}I))$. $\Tilde{g}$ is then used in place of $g$ to update the network parameters as in traditional SGD.

One of the biggest challenges with DP-SGD is tuning the clipping parameter $C$ since clipping greatly degrades the information stored in the original gradients~\cite{chen2020gs}. Choosing an optimal clipping value that does not significantly impact utility is crucial. However, tuning the clipping parameter is laborious as the optimal value fluctuates depending on network hyperparameters (i.e. model architecture, learning rate)~\cite{abadi2016deep}.
To avoid an intensive hyper-parameter search, \cite{chen2020gs}~proposes to use the Wasserstein loss with a gradient penalty term.  
This term ensures that the discriminator generates bounded gradient norms which are close to 1 under real and generated distributions. Therefore, an optimal clipping threshold of $C=1$ is obtained implicitly.

\algoplus trains the discriminator using differential private-SGD where the number of training iterations is determined based on the total privacy budget $(\epsilon$,$\delta)$. Thus, to compute the number of iterations, the privacy budget spent for every iteration must be bounded and accumulated. 
For this purpose we use the subsampled RDP analytical moments accountant technique. 

\begin{cor}
Each discriminator update satisfies $(\lambda,2B\lambda/\sigma^{2})$-RDP where $B$ is the batch size.
\end{cor}

\begin{prf}
Let $f=clip({\bar{g}_D},C)$ be the clipped gradient of the discriminator before adding noise. The sensitivity is derived via the triangle inequality:
\begin{equation}
    \Delta_{2}f = \max_{S,S'}||f(S)-f(S')||_{2} \leq 2C
\end{equation}

Since $C=1$ as a consequence of the Wasserstein loss with gradient penalty, 
and by using \eqref{eq:gaussian_mechanism}, the gaussian mechanism used within the DP-SGD procedure denoted as $\mathcal{M}_{\sigma}$ parameterized by noise scale $\sigma$ may be represented as being $(\lambda,2\lambda/\sigma^{2})$-RDP. 

Furthermore, each discriminator update for a batch of real data points $\{x_i,..,x_B\}$ can be represented as 

\begin{equation}
    \Tilde{g}_D = \frac{1}{B}\sum_{i=1}^{B}\mathcal{M}_{\sigma}(\triangledown_{\theta_D}\mathcal{L}_{D}(\theta_D,x_i))
\end{equation}

where $\tilde{g}_{D}$ and $\theta_{D}$ represent the perturbed gradients and the weights of the discriminator network, respectively. This may be regarded as a composition of $B$ Gaussian mechanisms and treated via \eqref{eq:composition}. The privacy cost for a single gradient update step for the discriminator can be expressed as $(\lambda,\sum_{i=1}^{B}2\lambda/\sigma^{2})$ or equivalently $(\lambda,2B\lambda/\sigma^{2})$.
\end{prf}

Note that $\mathcal{M}_{\sigma}$ is only applied for those gradients that are computed with respect to the real training dataset~\cite{abadi2016deep,zhang2018differentially}. Hence, the gradients computed with respect to the synthetic data and the gradient penalty term are left undisturbed. Next, to further amplify the privacy protection of the discriminator, we rely on~\eqref{eq:subsampling} with subsampling rate $\gamma=B/N$ where $B$ is the batch size and $N$ is the size of the training dataset. Intuitively, subsampling adds another layer of randomness and enhances privacy by decreasing the chances of leaking information about particular individuals who are not included in any given subsample of the dataset.

Lastly, it is worth mentioning that the \wgp training objective has one major pitfall with respect to the privacy cost. This is because, it encourages the use of a stronger discriminator network to provide more meaningful gradient updates to the generator. This requires performing multiple updates to the discriminator for each corresponding update to the generator leading to a faster consumption of the overall privacy budget. 

\color{black}

%% file: experiment_new.tex
\section{Experimental Analysis for Data utility}
\label{sec:experiment}
To show the efficacy of the proposed \algoplus, we select seven commonly used machine learning datasets, and compare with four SOTA GAN based tabular data generators and \algo. We evaluate the effectiveness of \algoplus in terms of the resulting ML utility, statistical similarity to the real data.  Moreover, we provide ablation analyses to highlight the efficacy of the unique components of \algoplus.

\subsection{Experimental Setup}
\label{ssec:setup}

{\bf Datasets}. Our algorithm is tested on seven commonly used machine learning datasets. Three of them    \textbf{Adult}
, \textbf{Covertype}
 and \textbf{Intrusion} 
 are from the UCI machine learning repository\footnote{\url{http://archive.ics.uci.edu/ml/datasets}}.
 \textbf{Credit} 
 and \textbf{Loan} 
 are from Kaggle\footnote{\url{https://www.kaggle.com/{mlg-ulb/creditcardfraud,itsmesunil/bank-loan-modelling}}}. The above five tabular datasets are used for classification tasks using as target a categorical \column. 
 To consider also regression tasks we use two more datasets, \textbf{Insurance}
 and \textbf{King} from Kaggle\footnote{\url{https://www.kaggle.com/{mirichoi0218/insurance,harlfoxem/housesalesprediction}}} where the target \column is continuous.
 
 Due to computing resource limitations, 50K rows of data are sampled randomly in a stratified manner with respect to the target \column for the Covertype, Credit and Intrusion datasets.  
The Adult, Loan, Insurance and King datasets are taken in their entirety. The details of each dataset are shown in Tab.~\ref{table:DD}. We assume that the data type of each \variable is known before training. \cite{ctgan} holds the same assumption.
\begin{table*}[t]
\centering
\caption{Description of Datasets}
\resizebox{1\textwidth}{!}{
\begin{tabular}{ |c|c|c|c|c|c|c|c|c|c| }
\hline
\textbf{Dataset} & \textbf{Problem}& \textbf{Train/Test Split} &\textbf{Target \column} & \textbf{$\mbox{Continuous}$}  & \textbf{$\mbox{Binary}$} & \textbf{$\mbox{Multi-class}$} & \textbf{$\mbox{Mixed-type}$}& \textbf{$\mbox{Long-tail}$}& \textbf{General Transform} \\
\hline
{Adult}  & Classification  & 39k/9k   & 'income'        & 3   & 2  & 7  & 2 & 0&1\\
\hline
  Covertype & Classification& 45k/5k    & 'Cover\_Type'    & 10   & 44 & 1  & 0 & 0&47\\\hline
 Credit    & Classification& 40k/10k    & 'Class'         & 30  & 1  & 0  & 0 & 1 &29\\\hline
  Intrusion & Classification& 45k/5k    & 'Class'         & 22   & 6  & 14 & 0 & 2 &6\\\hline
 Loan     & Classification & 4k/1k       & 'PersonalLoan'  & 5   & 5  & 2  & 1 & 0 &9\\\hline
 Insurance & Regression &1k/300&'charges'&13&3&2&2&0&1\\
 \hline
  King & Regression &17.3k/4.3k&'price'&1&6&2&0&0&6\\
\hline
\end{tabular}
}
\label{table:DD}
\end{table*}

{\bf Baselines}. Our \algoplus is compared with \algo and 4 other SOTA GAN-based tabular data generators: \ctgan, \tablegan, \cwgan and \medgan. To have a fair comparison, all algorithms are coded using Pytorch, with the generator and discriminator structures matching the descriptions provided in their respective papers. For Gaussian mixture estimation of continuous variables, we use the same settings as the evaluation of \ctgan, i.e. 10  modes. All algorithms are trained for 150 epochs for Adult, Covertype, Credit and Intrusion datasets, whereas the algorithms are trained for 300 epochs on Loan, Insurance and King datasets. The reason is these three datasets are smaller than the others and require more epochs to converge. 
Lastly, each experiment is repeated 3 times. 

{\bf Environment}. Experiments are run under Ubuntu 20.04 on a machine equipped with 32 GB memory, a GeForce RTX 2080 Ti GPU and a 10-core Intel i9 CPU.

\subsection{Evaluation Metrics}
\label{sec:metrics}
The evaluation is conducted on two dimensions: (1) machine learning (ML) utility, and (2) statistical similarity. They measure if the synthetic data can be used as a good proxy of the original data. 
  
\subsubsection{Machine Learning Utility}
\label{sec:ml_efficacy}
\begin{figure}[t]
    \centering
    \includegraphics[width=0.99\columnwidth]{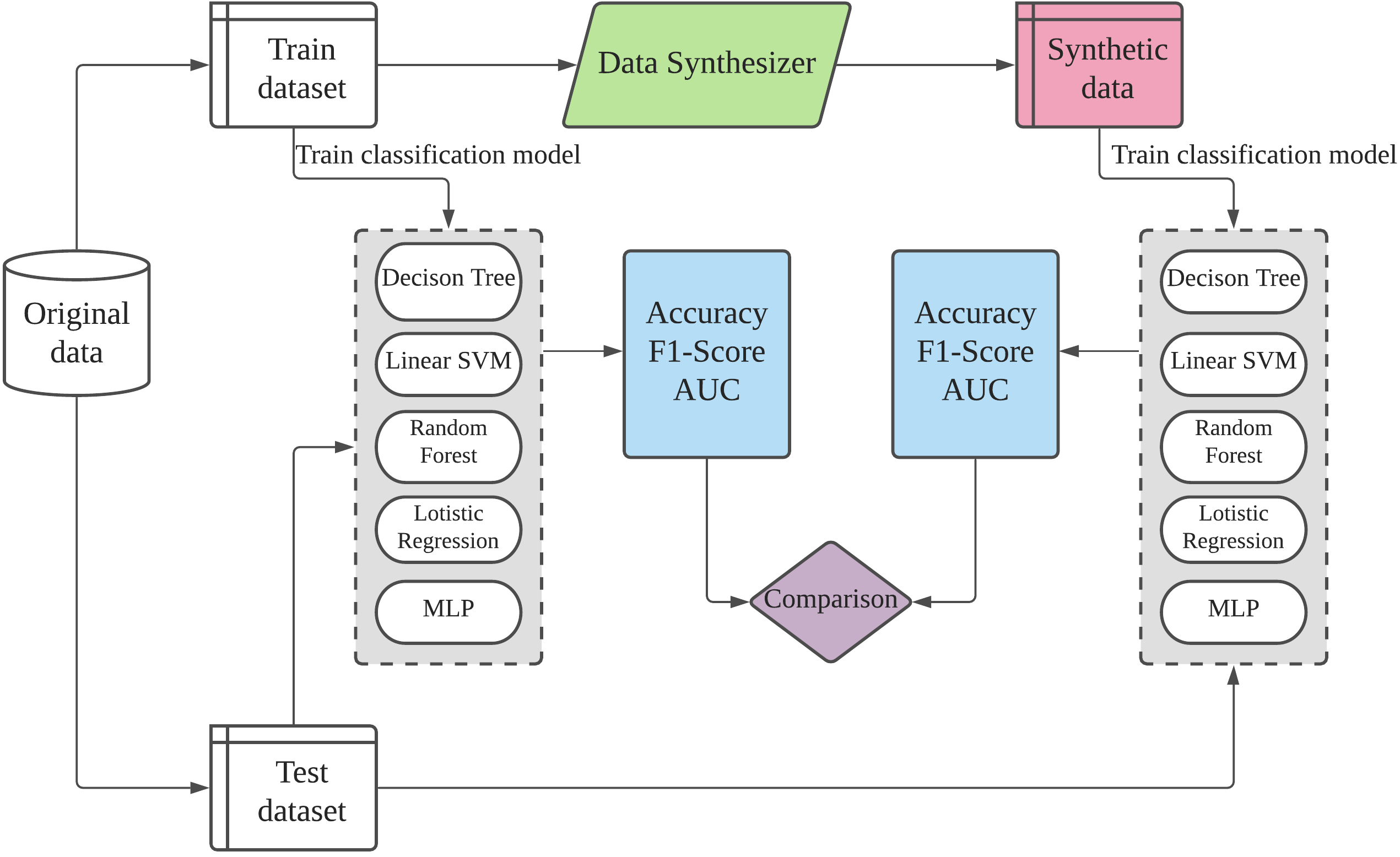}
    \caption{Evaluation flows for ML utility of Classification}
    \label{fig:settingA}
\end{figure}

The ML utility of classification and regression tasks is quantified differently. 
For classification, we quantify the ML utility via the performance, i.e, accuracy, F1-score and AUC, achieved by 5 widely used machine learning algorithms on real versus synthetic data: decision tree classifier, linear support-vector-machine (SVM), random forest classifier, multinomial logistic regression and MLP. Fig.~\ref{fig:settingA} shows the evaluation process for classification datasets. The training dataset and synthetic dataset are of the same size. The aim is to show the difference in ML utility when a ML model is trained on synthetic vs real data. We use different classification performance metrics. Accuracy is the most commonly used, but does not cope well with imbalanced target \columns. F1-score and AUC are more stable metrics for such cases. AUC ranges from 0 to 1.
For regression tasks, we quantify the ML utility in a similar manner but using 4 common regression algorithms -- linear regression, ridge regression, lasso regression and Bayesian ridge regression -- and 3  regression metrics -- mean absolute percentage error (MAPE), explained variance score (EVS) and $R^2$ score.
All algorithms are implemented using scikit-learn 0.24.2 with default parameters except max-depth 28 for decision tree and random forest, and 128 neurons for MLP.
For a fair comparison, hyper-parameters are fixed across all datasets. Due to this our results can slightly differ from~\cite{ctgan} where the authors use different ML models and hyper-parameters for different datasets.

\color{black}



\subsubsection{Statistical Similarity}
Three metrics are used to quantify the statistical similarity between real and synthetic data. 

\textbf{Jensen-Shannon divergence (JSD)}. The JSD provides a measure to quantify the difference between the probability mass distributions of individual categorical \columns belonging to the real and synthetic datasets, respectively. Moreover, this metric is bounded between 0 and 1 and is symmetric allowing for an easy interpretation of results. 

\textbf{Wasserstein distance (WD)}. In a similar vein, the Wasserstein distance is used to capture how well the distributions of individual continuous/mixed \columns are emulated by synthetically produced datasets in correspondence to real datasets. 
We use WD because we found that the JSD metric was numerically unstable for evaluating the quality of continuous \columns, especially when there is no overlap between the synthetic and original dataset. Hence, we resorted to utilize the more stable Wasserstein distance. 
 
\textbf{Difference in pair-wise correlation (Diff. Corr.)}. 
To evaluate how well feature interactions are preserved in the synthetic datasets, we first compute the pair-wise correlation matrix for the columns within real and synthetic datasets individually. 
Pearson correlation coefficient is used between any two continuous \columns. It ranges between $[-1,+1]$. Similarly, the Theil uncertainty coefficient is used to measure the correlation between any two categorical features. It ranges between $[0,1]$. And the correlation ratio  between categorical and continuous \columns is used. It also ranges between $[0,1]$. Note that the dython\footnote{\url{http://shakedzy.xyz/dython/modules/nominal/\#compute\_associations}} library is used to compute these metrics. Finally, the difference between pair-wise correlation matrices for real and synthetic datasets is computed.


\subsection{Results Analysis}

We first discuss the results in ML utility before addressing stochastic similarity.

\textbf{ML Utility}. Tab.~\ref{table:ML_all} shows the results for the classification datasets. A better synthetic dataset is expected to have small differences in d ML utility for classification tasks trained on real and synthetic data. It can be seen that \algoplus outperforms all other SOTA methods and \algo in all the metrics. 
\algoplus decreases the AUC difference from 0.094 (best baseline) to 0.041 (56.4\% reduction), and the difference in accuracy from 8.9\% (best baseline) to 5.23\% (41.2\% reduction). The improvement over \algo shows that general transform and \wgp loss indeed help enhance the feature representation and GAN training. Tab.~\ref{table:result_regression} shows the results for the regression datasets.
The result of \algo and \algoplus are far better than all other baselines. This shows the effectiveness of the feature engineering.  
Additionally, as \algoplus adds the auxiliary regressor which explicitly enhances the regression analysis, the overall downstream performance of \algoplus is better than \algo. We note that \algo uses auxiliary classification loss for the classification analysis and disables it for the regression analysis. 
\color{black}

\input{table1_alt}
\textbf{Statistical similarity}. Statistical similarity results for the classification datasets are reported in Tab.~\ref{table:ML_all} and for regression datasets in Tab.~\ref{table:result_regression}. \algoplus stands out again across all baselines in both groups of datasets. 
For classification datasets, \algoplus outperforms \algo, \ctgan and \tablegan by 37.1\%, 44.3\% and 51.3\% in average JSD. 
This is due to the use of the conditional vector, the log-frequency sampling and the extra losses, which work well for both balanced and imbalanced distributions. 
For continuous \columns (i.e. average WD),
the average WD column shows some extreme numbers such as 46257 and 238155 comparing to 484 of \algoplus. The reason is that these algorithms generate extremely large values for long tail \columns. Comparing to \algo, the significant improvement comes from the use of general transform to model continuous columns with simple distributions which originally used MSN under \algo and \ctgan. 
For regression datasets, \algoplus outperforms \algo by 63.4\% and 74.5\% in average JSD and average WD, respectively. Besides JSD and WD, the synthetic regression datasets maintain much better correlation than all the comparisons. This result confirms the efficacy of the usage of regressor.

 
\color{black} 


\input{table2_alt}

\subsection{Ablation Analysis}
\label{ssec:ablation}

For the sake of simplicity, ablation analysis are only implemented for classification datasets. 
We focus on conducting an ablation study to analyse the impact of the different components of \algo and \algoplus.

\subsubsection{With \algo}
\label{sssec:ctabgan}
To illustrate the efficiency of each strategy we implement four ablation studies which cut off the different components of \algo one by one: (1) \textbf{w/o $\mathcal{C}$}. In this experiment, Classifier $\mathcal{C}$ and the corresponding classification loss for Generator $\mathcal{G}$ are taken away from \algo; (2) \textbf{w/o I. loss} (information loss). In this experiment, we remove information loss from \algo; (3) \textbf{w/o MSN}. In this case, we substitute the mode specific normalization based on VGM for continuous \columns with min-max normalization and use simple one-hot encoding for categorical \columns. Here the conditional vector is the same as for \ctgan; (4) \textbf{w/o LT} (long tail). In this experiment, long tail treatment is no longer applied. This only affects datasets with long tailed columns, i.e. Credit and Intrusion.


The results are compared with the reference \algo implementing all strategies. 
All experiments are repeated 3 times, and results are evaluated on the same 5 machine learning algorithms introduced in Sec.~\ref{sec:ml_efficacy}. 
The test datasets and evaluation flow are the same as shown in Sec.~\ref{ssec:setup} and Sec.~\ref{sec:metrics}. 
Tab.~\ref{table:ablation_ctab} shows the results in terms of F1-score difference between ablation and \algo.
Each part of \algo has different impacts on different datasets. For instance, \textbf{w/o $\mathcal{C}$} has a negative impact for all datasets except Credit.
Since Credit has only 30 continuous \columns and one target \column, the semantic check can not be very effective.
\textbf{w/o information loss} has a positive impact for Loan, but results degenerate for all other datasets. It can even make the model unusable, e.g. for Intrusion. \textbf{w/o MSN} performs bad for Covertype, but has little impact for Intrusion. Credit w/o MSN performs  better than original \algo. This is because out of 30 continuous \columns, 28 are nearly single mode Gaussian distributed. The initialized high number of modes, i.e. 10, for each continuous \column (same setting as in \ctgan) degrades the estimation quality. \textbf{w/o LT} has the biggest impact on Intrusion, since it contains 2 long tail columns which are important predictors for the target column. For Credit, the influence is limited. Even if the long tail treatment fits well the \textit{amount} column (see Sec.~\ref{sec:further}), this variable is not a strong predictor for the target column.

\subsubsection{With \algoplus}
To show the efficacy of the General Transform and \wgp loss in \algoplus, we propose two ablation studies. (1)
\textbf{w/o GT} which disables the general transform in \algoplus. All continuous \columns use MSN and all the categorical \columns use one-hot encoding. (2) \textbf{w/o \wgp} which switches the default GAN training loss from \wgp to the original GAN loss defined in~\cite{gan}. It is worth noting that the information, \dsm and generator losses are still present in this experiment. The other experimental settings are the same as in Sec.~\ref{sssec:ctabgan}. Tab.~\ref{table:ablation_ctabganplus} shows the results in terms of F1-score difference among different versions of \algoplus. For Covertype, Credit and Intrusion datasets, the effects of GT and \wgp are all positive. GT significantly boosts the performance on Covertype and Credit datasets. But for Adult, it worsens the result. The reason is that the Adult dataset contains only one GT column: age. Since this column is strongly correlated with other columns, the original MSN encoding can better capture this interdependence. The positive impact of \wgp on the other hand is limited but consistent across all datasets. The only exception is the Loan dataset, where GT and \wgp have minor impacts. This is due to the fact that Loan has fewer \columns comparing to other datasets, which makes it easier to capture the correlation between columns. \algo already performs well on Loan, Therefore, GT and \wgp cannot further improve performance on this dataset.
 
\color{black}

\subsection{Results for Motivation Cases}
\label{sec:further}

\begin{figure*}[ht]
	\begin{center}
	
		\subfloat[\textit{bmi} in Insurrance]{
			\includegraphics[width=0.25\textwidth]{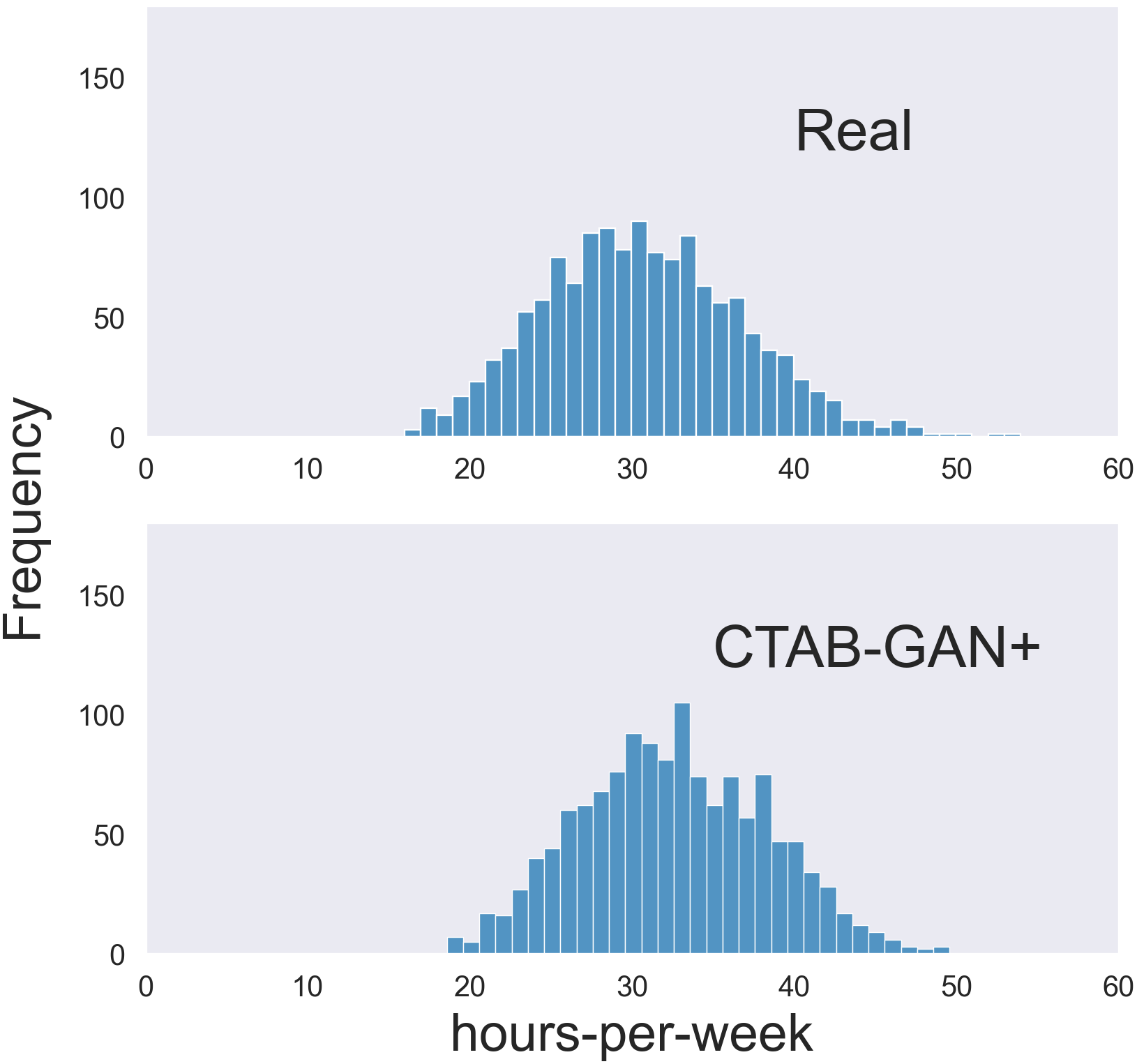}
			\label{fig:ctabgan_single_gaussian}
		}
		\subfloat[\textit{Mortgage} in Loan]{
			\includegraphics[width=0.25\textwidth]{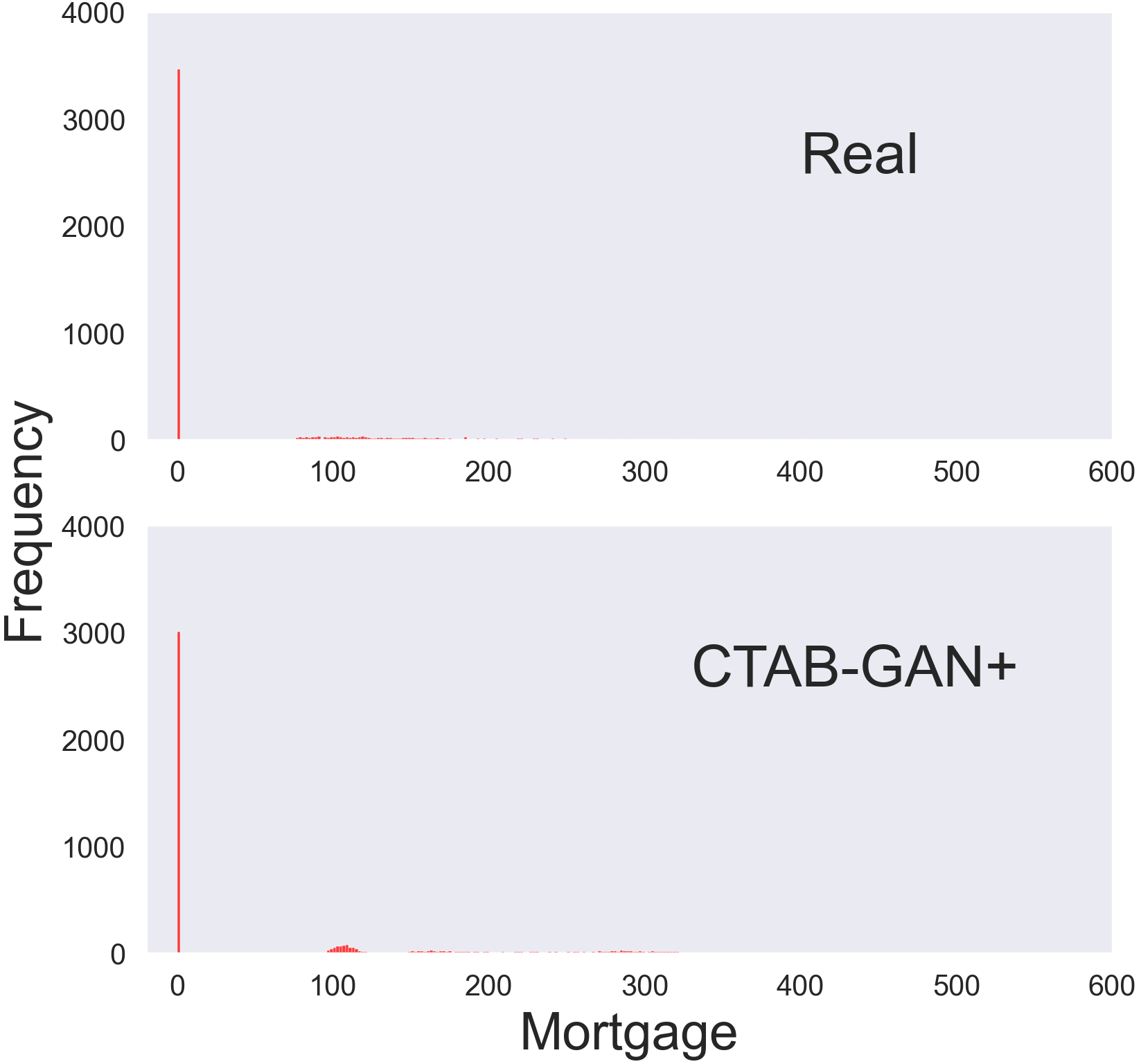}
			\label{fig:ctabgan_bimodal}
		}
		\subfloat[\textit{Amount} in Credit]{
			\includegraphics[width=0.22\textwidth]{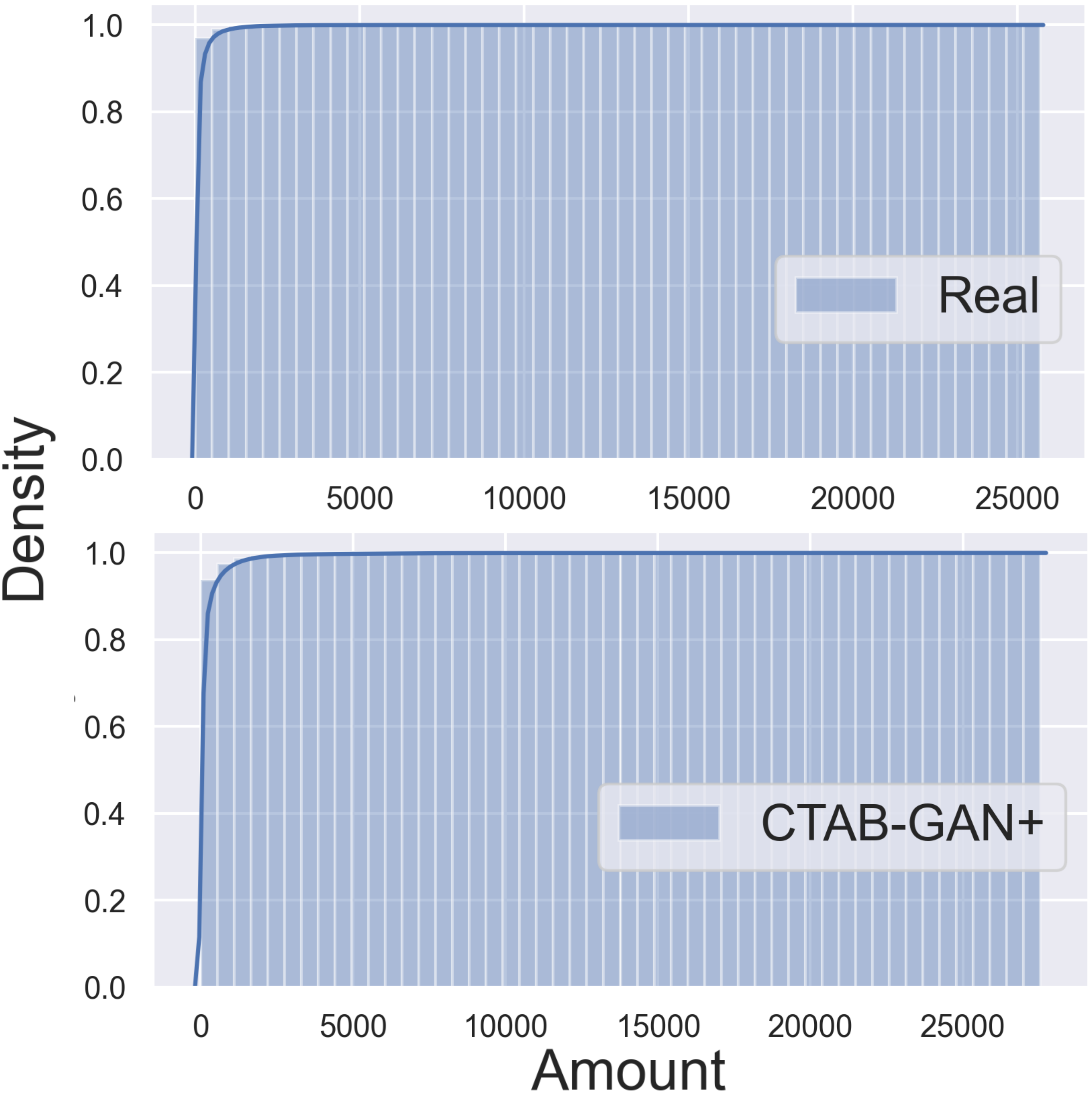}
			\label{fig:longtail_result}
		}
		\subfloat[\textit{Hours-per-week} in Adult]{
			\includegraphics[width=0.245\textwidth]{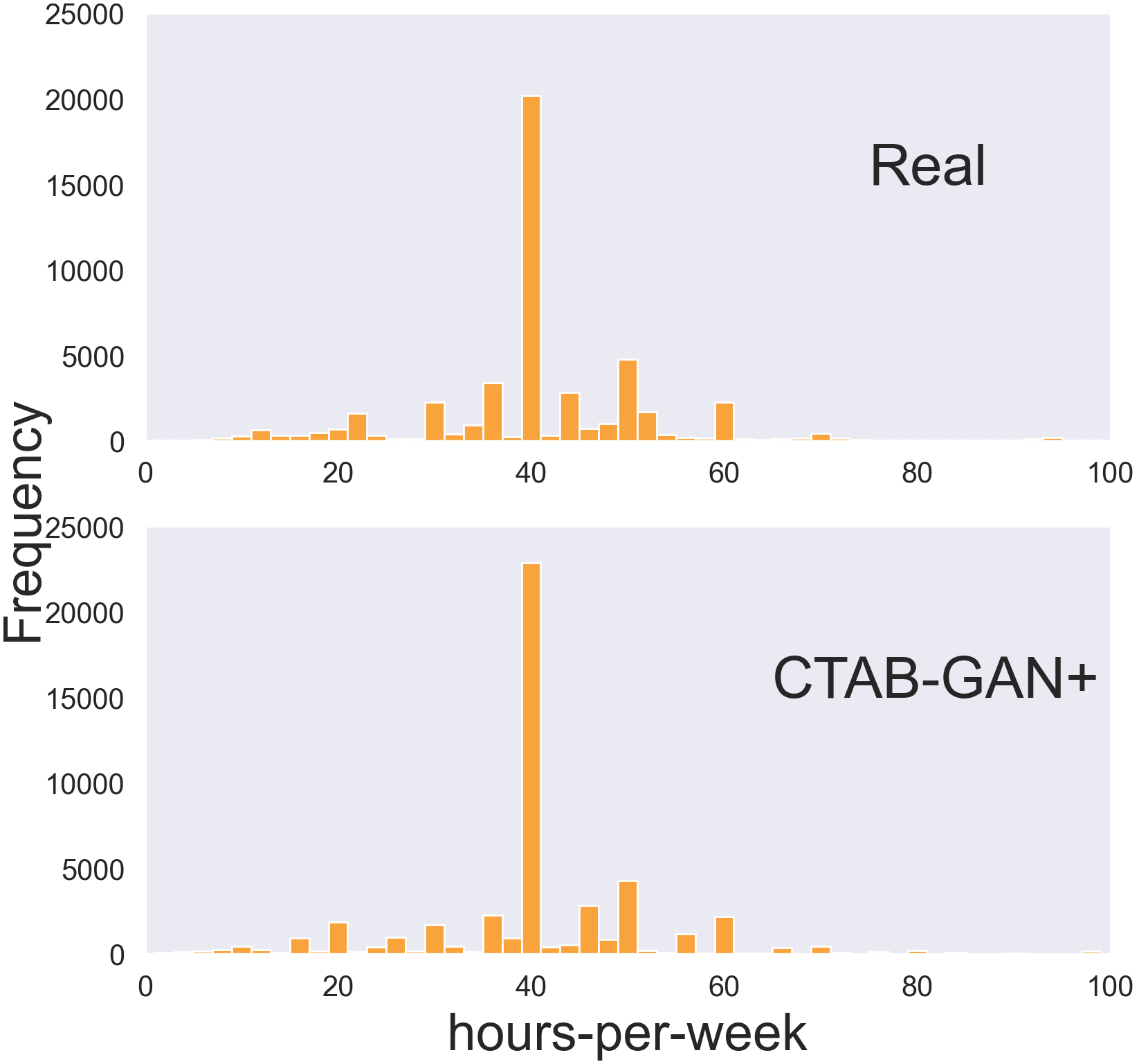}
			\label{fig:gmm_result}
		}
		\caption{ Challenges of modeling industrial dataset using existing GAN-based table generator: (a) simple gaussian (b) \bimodal type, (c) long tail distribution, and (d) \skewed data} 
		\label{fig:motivationcases_response}
 	\end{center}
\end{figure*}

After reviewing all the metrics, let us recall the four motivation cases from Sec.~\ref{sec:motivation}. 

{\bf Single Gaussian \variables}. Fig.~\ref{fig:ctabgan_single_gaussian}(a) shows the real and \algoplus generated \textit{bmi} \column.  \algoplus can reproduce the distribution with minor differences. This shows the effctiveness of general transform to better model \variables with single Gaussian distribution.

{\bf \Mixed data type \variables}. Fig.~\ref{fig:motivationcases_response}(b) compares the real and \algoplus generated \column \textit{Mortgage} in Loan dataset. \algoplus encodes this \column as \bimodal type. We can see that \algoplus generates clear 0 values and the frequency is close to real data. 

{\bf Long tail distributions.} Fig.~\ref{fig:motivationcases_response}(c) compares the cumulative frequency graph for the \textit{Amount} \column in Credit. This \column is a typical long tail distribution. 
One can see that \algoplus perfectly recovers the real distribution. Due to log-transform data pre-processsing, \algoplus learns this structure significantly better than the SOTA methods shown in Fig.~\ref{fig:motivationcases}(c).  

{\bf \Skewed multi-mode continuous \variables}.
Fig.~\ref{fig:motivationcases_response}(d) compares the frequency distribution for the continuous \column \textit{Hours-per-week} from Adult. Except the dominant peak at 40, there are many side peaks. Fig.~\ref{fig:motivationcases}(d), shows that \tablegan, \cwgan and \medgan struggle since they can learn only a simple Gaussian distribution due to the lack of any special treatment for continuous \columns. \ctgan, which also use VGM, can detect other modes. Still, \ctgan is not as good as \algo. The reason is that \ctgan lacks the mode of continuous \columns in the conditional vector. By incorporating the  mode of continuous \columns into conditional vector, we can apply the training-by-sample and logarithm frequency also to modes. This gives the mode with less weight more chance to appear in the training and avoids the mode collapse. 

%% file: table1_alt.tex
\begin{table}[t]
\centering
\caption{Difference of ML Utility and Statistical Similarity for Classification between original and synthetic data, averaged on five datasets}
\label{table:ML_all}
\resizebox{1.0\columnwidth}{!}{
\begin{tabular}{|c||c|c|c||c|c|c|}
\hline
\multirow{2}{*}{\textbf{Method}} & \multicolumn{3}{c||}{\bf ML Utility Difference} & \multicolumn{3}{c|}{\bf Statistical Similarity Difference}\\
\cline{2-7}
 & \textbf{Accuracy} & \textbf{F1-score}  & \textbf{AUC} & \textbf{Avg JSD} & \textbf{Avg WD}  & \textbf{Diff. Corr.}\\
\hline
\small{\algoplus}  & \textbf{5.23}\% & \textbf{0.090} & \textbf{0.041} & \textbf{0.039}  & \textbf{484}& \textbf{2.03}\\
\small{\algo}    & 8.90\% & 0.107 &   0.094  & 0.062  &  1197  &  2.09 \\
\small{CTGAN}    & 21.51\% & 0.274 &   0.253 & 0.070 &  1769  & 2.73   \\
\small{TableGAN} & 11.40\%  & 0.130 &   0.169 & 0.080 & 2117   &  2.30  \\
\small{MedGAN}    & 14.11\%&  0.282&   0.285 & 0.214 & 46257  &   5.48    \\
\small{CW-GAN}    &  20.06\%& 0.354 &  0.299 & 0.132 & 238155 &  5.82 \\
\hline
\end{tabular}
}
\end{table}

\begin{table}[t]
\centering
\caption{Difference of ML Utility and Statistical Similarity for Regression between original and synthetic data, averaged on two datasets
}
\label{table:result_regression}
\resizebox{1.0\columnwidth}{!}{
\begin{tabular}{|c||c|c|c||c|c|c|}
\hline
\multirow{2}{*}{\textbf{Method}} & \multicolumn{3}{c||}{\bf ML Utility Difference} & \multicolumn{3}{c|}{\bf Statistical Similarity Difference}\\
\cline{2-7}
 & \textbf{MAPE} & \textbf{EVS}  & \textbf{$R^2$} & \textbf{Avg JSD} & \textbf{Avg WD}  & \textbf{Diff. Corr.}\\
\hline
\small{\algoplus}  & \textbf{0.04} & \textbf{0.03} & \textbf{0.04} & \textbf{0.040}  & \textbf{856}& \textbf{0.65} \\
\small{\algo}    & 0.06& 0.05 &   0.06  & 0.119  &  3396  &  1.23\\
\small{CTGAN}   &0.87&0.59&0.71&0.139&3030&2.60\\
\small{TableGAN} &0.34  & 0.43 &   0.48 & 0.317 & 2366   &  2.26  \\
\small{MedGAN}    & 5.98&  0.65&   27.47 & 0.398 & 170307  &   7.27    \\
\small{CW-GAN}    &  0.64& 0.72 &  2.40 & 0.43 & 9.96E6 &  9.59 \\
\hline
\end{tabular}
}
\end{table}

%% file: table2_alt.tex
\begin{table}[htb]
\centering
\caption{Ablation Analysis For \algo (F1. diff.)}
\label{}
\resizebox{1.0\columnwidth}{!}{
\begin{tabular}{|c||c|c|c|c|c|}
\hline
\textbf{Dataset} & \textbf{\algo} & \textbf{w/o $\mathcal{C}$} & \textbf{w/o I. Loss}  & \textbf{w/o MSN} & \textbf{w/o LT} \\
\hline
Adult & 0.704 &  -0.01&	-0.037 & -0.05  & -  \\
Covertype  & 0.532 & -0.018 & -0.184& -0.118 & -   \\
Credit &  0.710 & +0.011& -0.177& +0.06 & +0.001   \\
Intrusion & 0.842&-0.031&-0.437&+0.003 & -0.074  \\
Loan & 0.803 &-0.044&+0.028&+0.013 & -  \\
\hline
\end{tabular}
}
\label{table:ablation_ctab}
\end{table}

\begin{table}
    \centering
    \caption{Ablation Analysis For \algoplus (F1. diff.)}
    \medskip
    \resizebox{1\columnwidth}{!}{
\begin{tabular}{ |C{1.8cm}||C{1.8cm}|C{1.8cm}|C{2.2cm}|}
\hline
\textbf{Dataset} & \textbf{\algoplus}& \textbf{w/o GT} & \textbf{w/o \wgp}  \\
\hline
Adult &0.684&+0.013&-0.029 \\
Covertype &0.636&-0.196&-0.012\\
Credit &0.802&-0.303&-0.08 \\
Intrusion &0.912&-0.041&-0.049 \\
Loan &0.806&-0.001&+0.003 \\
\hline
\end{tabular}
}
\label{table:ablation_ctabganplus}
\end{table}

%% file: experiment_dp.tex
\section{Experiment Analysis for Differential Privacy}
\label{sec:dp}
In this section, we show the effect of adding DP to \algoplus and compare \algoplus with three SOTA DP GAN algorithms.

\subsection{Experiment Setup}
{\bf Datasets}. For sake of simplicity, we only use the classification datasets: Adult, Covertype, Intrusion, Credit and Loan.

{\bf Metrics}. We use the same ML utility metrics from Section \ref{sec:metrics} under two privacy budgets, i.e., $\epsilon = 1$ and $\epsilon = 100$.  

{\bf Baselines}. \algoplus is compared against 3 SOTA architectures: PATE-GAN~\cite{pategan}, DP-WGAN~\cite{xie2018differentially} and GS-WGAN~\cite{chen2020gs}. The code of PATE-GAN and DP-WGAN is taken from Private Data Generation Toolbox\footnote{\url{https://github.com/BorealisAI/private-data-generation}}  
which already adapts them for tabular data synthesis. We extend GS-WGAN to the tabular domain by converting each data row into a bitmap image. We first normalize all values to the range $[0,1]$ and re-shape rows in the form of square images filling missing entries (if any) with zeros. The re-shaped rows are fed into the algorithm and the generated images are transformed into data rows by reversing the previous two operations. 
All hyper-parameters are kept to their default values except for the default network architecture which is adjusted according to the spatial dimensions of the tabular datasets. Lastly, note that to compute privacy cost fairly, the RDP accountant is used for all approaches that use DP-SGD as it provides tighter privacy guarantees than the moment accountant~\cite{wang2019subsampled}.

{\bf Privacy accounting}. 
To compute the privacy cost in a fair manner, we use the RDP accountant for all approaches that employ DP-SGD: \algoplus, DP-WGAN and GS-WGAN. PATE-GAN uses moment accountant~\cite{wang2019subsampled} by default. We set $\delta=10^{-5}$ for all experiments.
We follow the examples of DP-WGAN and set the exploration span of $\lambda$ to $[2, 4096]$.
We use~\eqref{eq:rdp-to-dp} to convert the overall cumulative privacy cost computed in terms of RDP back to $(\epsilon,\delta)$-DP.

\subsection{Results Analysis}

{\bf ML Utility}
Tab.~\ref{table:ML_allE} presents the results for the differences ML utility between models trained on the original and synthetic data: lower is better. 
\algoplus outperforms all other SOTA algorithms under both privacy budgets. 
With a looser privacy budget, i.e., higher $\epsilon$, almost all  metrics for all algorithms improve. The only exception is AUC for GS-WGAN, but the difference is minor. These results are in line with our expectation because higher privacy budgets mean training the model with less injected noise and more training epochs -- before exhaustion of the privacy budget. The superior performance of \algoplus compared to other baselines can be explained by its sophisticated neural network architecture, i.e., conditional GAN, which improves the training objective and capacity to better deal with the challenges of the tabular domain such as imbalanced categorical columns and mixed data-types. This also explains the poor results offered by GS-WGAN which is not designed to handling these specific issues achieving the worst overall performance.

\input{table_dp}

{\bf Statistical Similarity}.
Tab.~\ref{table:SS_allE} summarizes the statistical similarity results.
Among all DP models, \algoplus and GS-WGAN consistently improve across all metrics when the privacy budget is increased. But the performance of GS-WGAN is significantly worse than \algoplus. With a higher privacy budget, Avg WD of PATE-GAN is slightly increased. And the correlation difference of DP-WGAN increases too. This highlights the inability of this methods to capture the statistical distributions during training despite a looser privacy budget. This can be explained by the lack of an effective training framework for dealing with complex statistical distributions present in the tabular domain which arise from imbalances in categorical columns and skews in continuous columns commonly found in real-world tabular datasets.


\color{black}

%% file: table_dp.tex
\begin{table}[t]
\centering
\caption{Difference of accuracy (\%), F1-score, AUC and AP between original and synthetic data: average over 5 ML models and 5 datasets with different privacy budgets $\epsilon=1$ $\&$ $\epsilon=100$.}
\resizebox{\columnwidth}{!}{
\begin{tabular}{|c||c|c|c||c|c|c|}
\hline
\multirow{2}{*}{\textbf{Method}} & \multicolumn{3}{c||}{\bf $\epsilon=1$} &
\multicolumn{3}{c|}{\bf $\epsilon=100$} \\
\cline{2-7}
 & \textbf{Accuracy} & \textbf{F1-Score} & \textbf{AUC}  & \textbf{Accuracy}  & \textbf{F1-Score} & \textbf{AUC}\\
\hline
{\algoplus}  &\textbf{19.12\%}  &\textbf{0.320}&\textbf{0.311}& \textbf{13.34}\%&\textbf{0.311} &\textbf{0.299}\\
{PATE-GAN}   &38.17\%  &0.513&0.447&37.96\% &0.508&0.394 \\
{DP-WGAN}    &36.88\%  &0.536&0.510&27.27\% &0.502&0.423 \\
{GS-WGAN}    &64.10\%  &0.668&0.492&58.22\% &0.639&0.498 \\
\hline
\end{tabular}
}
\label{table:ML_allE}
\end{table}

\begin{table}[t]
\centering 
\caption{Statistical similarity metrics between original and synthetic data: average on 5 datasets with different privacy budgets $\epsilon=1$ $\&$ $\epsilon=100$.}
\resizebox{\columnwidth}{!}{
\centering
\begin{tabular}{|c||c|c|c||c|c|c|}
\hline
\multirow{2}{*}{\textbf{Method}} & \multicolumn{3}{c||}{\bf $\epsilon=1$} &
\multicolumn{3}{c|}{\bf $\epsilon=100$} \\
\cline{2-7}
 & \textbf{Avg JSD} & \textbf{Avg WD} & \textbf{Diff. Corr.} & 
\textbf{Avg JSD} & \textbf{Avg WD} & \textbf{Diff. Corr.} \\

\hline
{\algoplus}    &\textbf{0.192}&\textbf{775}&\textbf{5.61}&\textbf{0.137}&\textbf{655}&\textbf{5.54}\\
{PATE-GAN}     &0.356&8632&9.45&0.366&8634&8.94\\
{DP-WGAN}      &0.362&8632&9.19&0.359&8632&9.49\\
{GS-WGAN} &0.624&4.07E+06&15.61&0.547&54574&13.74\\
\hline
\end{tabular}
}
\label{table:SS_allE}
\end{table}

%% file: main.bbl
\begin{thebibliography}{10}

\bibitem{abadi2016deep}
M.~Abadi, A.~Chu, I.~Goodfellow, H.~B. McMahan, I.~Mironov, K.~Talwar, and
  L.~Zhang.
\newblock Deep learning with differential privacy.
\newblock In {\em ACM SIGSAC Conference on Computer and Communications Security
  (CCS)}, 2016.

\bibitem{wgan}
M.~Arjovsky, S.~Chintala, and L.~Bottou.
\newblock Wasserstein generative adversarial networks.
\newblock In {\em Proceedings of the 34th ICML - Volume 70}, page 214–223.
  JMLR.org, 2017.

\bibitem{pmlr-v70-arjovsky17a}
M.~Arjovsky, S.~Chintala, and L.~Bottou.
\newblock {W}asserstein generative adversarial networks.
\newblock In {\em International Conference on Machine Learning (ICML)}, 2017.

\bibitem{cramerdistance}
M.~G. Bellemare, I.~Danihelka, W.~Dabney, S.~Mohamed, B.~Lakshminarayanan,
  S.~Hoyer, and R.~Munos.
\newblock The cramer distance as a solution to biased wasserstein gradients.
\newblock {\em ArXiv}, abs/1705.10743, 2017.

\bibitem{prml}
C.~M. Bishop.
\newblock {\em Pattern Recognition and Machine Learning (Information Science
  and Statistics)}.
\newblock Springer-Verlag, Berlin, Heidelberg, 2006.

\bibitem{chen2020gs}
D.~Chen, T.~Orekondy, and M.~Fritz.
\newblock Gs-wgan: A gradient-sanitized approach for learning differentially
  private generators.
\newblock {\em arXiv preprint arXiv:2006.08265}, 2020.

\bibitem{gan_leak}
D.~Chen, N.~Yu, Y.~Zhang, and M.~Fritz.
\newblock Gan-leaks: A taxonomy of membership inference attacks against
  generative models.
\newblock In {\em ACM SIGSAC Conference on Computer and Communications Security
  (CCS)}, 2020.

\bibitem{medgan}
E.~Choi, S.~Biswal, B.~Malin, J.~Duke, W.~F. Stewart, and J.~Sun.
\newblock Generating multi-label discrete patient records using generative
  adversarial networks.
\newblock {\em arXiv preprint arXiv:1703.06490}, 2017.

\bibitem{dwork2008differential}
C.~Dwork.
\newblock Differential privacy: A survey of results.
\newblock In {\em International Conference on Theory and Applications of Models
  of Computation (TAMC)}. Springer, 2008.

\bibitem{dwork2014algorithmic}
C.~Dwork, A.~Roth, et~al.
\newblock The algorithmic foundations of differential privacy.
\newblock {\em Foundations and Trends in Theoretical Computer Science}, 2014.

\bibitem{cwgan}
J.~Engelmann and S.~Lessmann.
\newblock Conditional wasserstein gan-based oversampling of tabular data for
  imbalanced learning.
\newblock {\em arXiv preprint arXiv:2008.09202}, 2020.

\bibitem{gan}
I.~J. Goodfellow, J.~Pouget-Abadie, M.~Mirza, B.~Xu, D.~Warde-Farley, S.~Ozair,
  A.~Courville, and Y.~Bengio.
\newblock Generative adversarial nets.
\newblock In {\em Proceedings of the 27th NIPS - Volume 2}, page 2672–2680,
  Cambridge, MA, USA, 2014.

\bibitem{wgan_gp}
I.~Gulrajani, F.~Ahmed, M.~Arjovsky, V.~Dumoulin, and A.~Courville.
\newblock Improved training of wasserstein gans.
\newblock In {\em the 31st NIPS}, page 5769–5779, 2017.

\bibitem{gulrajani2017improved}
I.~Gulrajani, F.~Ahmed, M.~Arjovsky, V.~Dumoulin, and A.~C. Courville.
\newblock Improved training of wasserstein gans.
\newblock In {\em Advances in Neural Information Processing Systems}, 2017.

\bibitem{Hawes2020Implementing}
M.~B. Hawes.
\newblock Implementing differential privacy: Seven lessons from the 2020 united
  states census.
\newblock {\em Harvard Data Science Review}, 4 2020.

\bibitem{pategan}
J.~Jordon, J.~Yoon, and M.~van~der Schaar.
\newblock Pate-gan: Generating synthetic data with differential privacy
  guarantees.
\newblock In {\em International Conference on Learning Representations (ICLR)},
  2018.

\bibitem{stylegan}
T.~{Karras}, S.~{Laine}, and T.~{Aila}.
\newblock A style-based generator architecture for generative adversarial
  networks.
\newblock In {\em IEEE/CVF CVPR}, pages 4396--4405, 2019.

\bibitem{pacgan}
Z.~{Lin}, A.~{Khetan}, G.~{Fanti}, and S.~{Oh}.
\newblock Pacgan: The power of two samples in generative adversarial networks.
\newblock {\em IEEE JSAIT}, 1(1):324--335, 2020.

\bibitem{long2019scalable}
Y.~Long, S.~Lin, Z.~Yang, C.~A. Gunter, and B.~Li.
\newblock Scalable differentially private generative student model via pate.
\newblock {\em arXiv preprint arXiv:1906.09338}, 2019.

\bibitem{mironov2017renyi}
I.~Mironov.
\newblock R{\'e}nyi differential privacy.
\newblock In {\em Computer Security Foundations Symposium (CSF)}. IEEE, 2017.

\bibitem{cramergan}
A.~{Mottini}, A.~{Lheritier}, and R.~{Acuna-Agost}.
\newblock {Airline Passenger Name Record Generation using Generative
  Adversarial Networks}.
\newblock In {\em workshop on Theoretical Foundations and Applications of Deep
  Generative Models. ICML}, July 2018.

\bibitem{narayanan2008}
A.~{Narayanan} and V.~{Shmatikov}.
\newblock Robust de-anonymization of large sparse datasets.
\newblock In {\em IEEE Symposium on Security and Privacy}, pages 111--125,
  2008.

\bibitem{acgan}
A.~Odena, C.~Olah, and J.~Shlens.
\newblock Conditional image synthesis with auxiliary classifier gans.
\newblock In {\em The 34th ICML - Volume 70}, page 2642–2651, 2017.

\bibitem{papernot2016semi}
N.~Papernot, M.~Abadi, U.~Erlingsson, I.~Goodfellow, and K.~Talwar.
\newblock Semi-supervised knowledge transfer for deep learning from private
  training data.
\newblock {\em arXiv preprint arXiv:1610.05755}, 2016.

\bibitem{tablegan}
N.~Park, M.~Mohammadi, K.~Gorde, S.~Jajodia, H.~Park, and Y.~Kim.
\newblock Data synthesis based on generative adversarial networks.
\newblock {\em Proc. VLDB Endow.}, 11(10):1071–1083, June 2018.

\bibitem{proven2021comicgan}
B.~Proven-Bessel, Z.~Zhao, and L.~Chen.
\newblock Comicgan: Text-to-comic generative adversarial network.
\newblock {\em arXiv preprint arXiv:2109.09120}, 2021.

\bibitem{priv_mirage}
T.~Stadler, B.~Oprisanu, and C.~Troncoso.
\newblock Synthetic data--a privacy mirage.
\newblock {\em arXiv preprint arXiv:2011.07018}, 2020.

\bibitem{torfi2020differentially}
A.~Torfi, E.~A. Fox, and C.~K. Reddy.
\newblock Differentially private synthetic medical data generation using
  convolutional gans.
\newblock {\em arXiv preprint arXiv:2012.11774}, 2020.

\bibitem{torkzadehmahani2019dp}
R.~Torkzadehmahani, P.~Kairouz, and B.~Paten.
\newblock Dp-cgan: Differentially private synthetic data and label generation.
\newblock In {\em Conference on Computer Vision and Pattern Recognition (CVPR)
  Workshops}, 2019.

\bibitem{crossnet}
R.~Wang, B.~Fu, G.~Fu, and M.~Wang.
\newblock Deep \& cross network for ad click predictions.
\newblock In {\em Proceedings of the ADKDD'17}, New York, NY, USA, 2017.

\bibitem{wang2019subsampled}
Y.-X. Wang, B.~Balle, and S.~P. Kasiviswanathan.
\newblock Subsampled r{\'e}nyi differential privacy and analytical moments
  accountant.
\newblock In {\em International Conference on Artificial Intelligence and
  Statistics (AISTATS)}, 2019.

\bibitem{xie2018differentially}
L.~Xie, K.~Lin, S.~Wang, F.~Wang, and J.~Zhou.
\newblock Differentially private generative adversarial network.
\newblock {\em arXiv preprint arXiv:1802.06739}, 2018.

\bibitem{ctgan}
L.~Xu, M.~Skoularidou, A.~Cuesta-Infante, and K.~Veeramachaneni.
\newblock Modeling tabular data using conditional gan.
\newblock In {\em NIPS}, 2019.

\bibitem{yahi_gan}
A.~Yahi, R.~Vanguri, and N.~Elhadad.
\newblock Generative adversarial networks for electronic health records: A
  framework for exploring and evaluating methods for predicting drug-induced
  laboratory test trajectories.
\newblock In {\em NIPS workshop}, 2017.

\bibitem{younesian2020qactor}
T.~Younesian, Z.~Zhao, A.~Ghiassi, R.~Birke, and L.~Y. Chen.
\newblock Qactor: On-line active learning for noisy labeled stream data.
\newblock {\em arXiv preprint arXiv:2001.10399}, 2020.

\bibitem{zhang2018differentially}
X.~Zhang, S.~Ji, and T.~Wang.
\newblock Differentially private releasing via deep generative model (technical
  report).
\newblock {\em arXiv preprint arXiv:1801.01594}, 2018.

\bibitem{Zhao:TDSC21}
Z.~Zhao, R.~Birke, R.~Han, B.~Robu, S.~Bouchenak, S.~B. Mokhtar, and L.~Y.
  Chen.
\newblock Enhancing robustness of on-line learning models on highly noisy data.
\newblock {\em IEEE Transactions on Dependable and Secure Computing},
  18(5):2177--2192, 2021.

\bibitem{Zhao:DSN19}
Z.~Zhao, S.~Cerf, R.~Birke, B.~Robu, S.~Bouchenak, S.~B. Mokhtar, and L.~Y.
  Chen.
\newblock Robust anomaly detection on unreliable data.
\newblock In {\em 49th Annual {IEEE/IFIP} International Conference on
  Dependable Systems and Networks}, 2019.

\bibitem{ctabgan}
Z.~Zhao, A.~Kunar, R.~Birke, and L.~Y. Chen.
\newblock Ctab-gan: Effective table data synthesizing.
\newblock In {\em Proceedings of The 13th Asian Conference on Machine
  Learning}, volume 157, pages 97--112, 17--19 Nov 2021.

\end{thebibliography}
